\documentclass[final]{cvpr}

\usepackage{times}
\usepackage{epsfig}
\usepackage{graphicx}
\usepackage{amsmath}
\usepackage{amssymb}

\usepackage{color}
\usepackage{kotex}
\usepackage{cases}
\usepackage{nicefrac}
\usepackage{relsize}
\usepackage{soul}  
\usepackage{multirow}
\usepackage{caption}
\usepackage{soul}
\usepackage{helvet}  
\usepackage{courier}
\usepackage{url} 
\usepackage{makecell}
\usepackage{enumitem}
\usepackage{subcaption}
\usepackage{tabularx}
\usepackage{booktabs}
\usepackage{tablefootnote}
\usepackage{scrextend}
\newcolumntype{Y}{>{\centering\arraybackslash}X}

\usepackage[pagebackref=true,breaklinks=true,colorlinks,bookmarks=false]{hyperref}

\def\confYear{CVPR 2021}
\begin{document}
\thispagestyle{empty}
\title{Unsupervised Hyperbolic Representation Learning \\via Message Passing Auto-Encoders}

\author{Jiwoong Park\thanks{equally contributed.} $^1$ \quad Junho Cho\footnotemark[1] $^{1}$ \quad Hyung Jin Chang$^2$ \quad Jin Young Choi$^1$\\
\vspace{-1mm}
{\small $^1$ASRI, Dept. of ECE., Seoul National University}
{\small \hspace{1cm}$^2$School of Computer Science, University of Birmingham} \\
{\tt\scriptsize \{ptywoong,junhocho,jychoi\}@snu.ac.kr, h.j.chang@bham.ac.uk}}

\maketitle

\begin{abstract}
Most of the existing literature regarding hyperbolic embedding concentrate upon supervised learning, whereas the use of unsupervised hyperbolic embedding is less well explored.
In this paper, we analyze how unsupervised tasks can benefit from learned representations in hyperbolic space. 
To explore how well the hierarchical structure of unlabeled data can be represented in hyperbolic spaces, we design a novel hyperbolic message passing auto-encoder whose overall auto-encoding is performed in hyperbolic space.
The proposed model conducts auto-encoding the networks via fully utilizing hyperbolic geometry in message passing.
Through extensive quantitative and qualitative analyses, we validate the properties and benefits of the unsupervised hyperbolic representations.
Codes are available at \url{https://github.com/junhocho/HGCAE}.
\end{abstract}

\section{Introduction}
A fundamental problem of machine learning is learning useful representations from high-dimensional data.
There are many supervised representation learning methods which achieve good performances for downstream tasks \cite{krizhevsky2012imagenet, kipf2017semi, long2015fully, yan2018spatial} on several data domains such as images and graphs.
In recent years, with the success of deep learning, various large-scale real-world datasets have been collated \cite{krizhevsky2012imagenet, krizhevsky2009learning, wah2011caltech, sen2008collective}. 
However, the larger these datasets and the closer they are to the real world, the expense and effort required to label the data increases proportionally.
Thus, unsupervised representation learning is an increasingly viable approach to extract useful representation from real-world datasets.

Recently, many works \cite{nickel2017poincare, nickel2018learning, ganea2018hyperbolic, gulcehre2018hyperbolic, chami2019hyperbolic, bachmann2019constant, khrulkov2020hyperbolic} utilize hyperbolic geometry \cite{krioukov2010hyperbolic} to learn representations by understanding the underlying nature of the data domains.
It is well known that complex networks contain latent hierarchies between large groups and the divided subgroups of nodes, and can be approximated as trees that grow exponentially with their depth \cite{krioukov2010hyperbolic}. 
Based on this fact, previous works which involve graphs \cite{bouchard2015approximate, nickel2014reducing, nickel2017poincare, nickel2018learning, ganea2018hyperbolic, nagano2019wrapped} showed the effectiveness of learning representation using hyperbolic spaces (a continuous version of trees) where distances increase exponentially when moving away from the origin.
More recently, works \cite{chami2019hyperbolic, liu2019hyperbolic, bachmann2019constant} have been conducted which learn more powerful representations via conducting message passing (graph convolution) \cite{gilmer2017neural, kipf2017semi, velickovic2018graph} in hyperbolic spaces.

In addition, it has been successfully shown that grafting hyperbolic geometry onto computer vision tasks is promising \cite{khrulkov2020hyperbolic}.
They observed a high degree of hyperbolicity \cite{fournier2015computing} in the activations of image datasets obtained from pre-trained convolutional networks.
Also, it has been shown that the hyperbolic distance between learned embeddings and the origin of the Poincar\'e ball could be considered as a measurement of the model's confidence.
Using these analyses, \cite{khrulkov2020hyperbolic} added a single layer of hyperbolic neural networks \cite{ganea2018hyperbolic} to deep convolutional networks and showed the benefits of hyperbolic embeddings on few-shot learning and person re-identification.
Another work \cite{liu2020hyperbolic} also demonstrated the suitability of hyperbolic embeddings on zero-shot learning.
However, most of the existing hyperbolic representation learning works \cite{khrulkov2020hyperbolic, liu2020hyperbolic, chami2019hyperbolic, liu2019hyperbolic, bachmann2019constant} mainly focus on a supervised setting, and the effect of hyperbolic geometry on unsupervised representation learning has not been explored deeply so far \cite{mathieu2019poincare, grattarola2019adversarial, nagano2019wrapped}.

In this paper, we explore the benefits of hyperbolic geometry to carry out unsupervised representation learning upon various data domains.
Our motivation is to learn high-quality node embeddings of the graphs that are hierarchical and tree-like without supervision via considering the geometry of the embedding space.
To do so, we present a novel hyperbolic graph convolutional auto-encoder (HGCAE) by combining hyperbolic geometry and message passing \cite{gilmer2017neural}. 
Every layer of HGCAE performs message passing in the hyperbolic space and its corresponding tangent space where curvature values can be trained. 
This is primarily in contrast to the Poincar\'e variational auto-encoder (P-VAE) \cite{mathieu2019poincare} whose latent space is the Poincar\'e ball and conducts message passing in Euclidean space.
The HGCAE conducts auto-encoding the graphs from diverse data domains, such as images or social networks, in the hyperbolic space such as the Poincar\'e ball and hyperboloid.
To fully utilize hyperbolic geometry for representation learning, we adopt a geometry-aware attention mechanism \cite{gulcehre2018hyperbolic} when conducting message passing.
Through extensive experiments and analyses using the learned representation in the hyperbolic latent spaces, we present the following observations on hierarchically structured data:

\begin{itemize}[leftmargin=*]
\item The proposed auto-encoder, which combines message passing based on geometry-aware attention and hyperbolic spaces, can learn useful representations for downstream tasks.
On various networks,
the proposed method achieves state-of-the-art results on node clustering and link prediction tasks.
\item Image clustering tasks can benefit from embeddings in hyperbolic latent spaces. 
We achieve comparable results to state-of-the-art image clustering results by learning representations from the activations of neural networks. 
\item Hyperbolic embeddings of images, the results of unsupervised learning, can recognize the underlying data structures such as a class hierarchy without any supervision of ground-truth class hierarchy.
\item We show that the sample's hyperbolic distance from the origin in hyperbolic space can be utilized as a criterion to choose samples, therefore improving the generalization ability of a model for a given dataset.
\end{itemize}

\section{Related Works}
\noindent \textbf{Hyperbolic embedding of images.} Khrulkov et al. \cite{khrulkov2020hyperbolic} validated hyperbolic embeddings of images via measuring the degree of hyperbolicity of image datasets.
Many datasets such as CIFAR10/100 \cite{krizhevsky2009learning}, CUB \cite{wah2011caltech} and MiniImageNet \cite{ravi2016optimization} showed high degrees of hyperbolicity. 
In particular, the ImageNet dataset \cite{russakovsky2015imagenet} is organized by following the hierarchical structure of WordNet \cite{miller1998wordnet}.
These observations suggest that hyperbolic geometry can be beneficial in analyzing image manifolds by capturing not only semantic similarities but also hierarchical relationships between images.
Furthermore, Khrulkov et al. \cite{khrulkov2020hyperbolic} empirically showed that the distance between the origin and the image embeddings in the Poincar\'e ball can be regarded as the measure of model's confidence.
They observed that the samples which are easily classified are located near the boundary while those more ambiguous samples lie near the origin of the hyperbolic space.
Recent works of hyperbolic image embeddings \cite{khrulkov2020hyperbolic, liu2020hyperbolic} add one or two layers of hyperbolic layers \cite{ganea2018hyperbolic} after an Euclidean convolutional network. 

\noindent \textbf{Graph auto-encoding via hyperbolic geometry.} Some recent works \cite{grattarola2019adversarial, mathieu2019poincare, skopek2019mixed} attempted to auto-encode graphs in hyperbolic space.
Their models attempted to learn latent representations in the hyperbolic space via grafting hyperbolic geometry onto a variational auto-encoder model \cite{kingma2013auto}. 
\cite{grattarola2019adversarial, mathieu2019poincare} encoded the node representation via message passing \cite{kipf2017semi} in Euclidean space, then the encoded representation was projected onto the hyperbolic space.
Similar to these concurrent models, our auto-encoder framework learns latent node representations of the graph in hyperbolic latent spaces.
Differing from these models, our work considers hyperbolic geometry throughout the auto-encoding process.
Each encoder and decoder layer of the proposed model conducts message passing by utilizing geometry-aware attention in the hyperbolic space and its tangent space.

\section{Hyperbolic Geometry}
A real, smooth manifold $\mathcal{M}$ is a set of points $x$, that is locally similar to linear space.
At each point $x \in \mathcal{M}$, the tangent space at $x$, $\mathcal{T}_x\mathcal{M}$, is a real vector space whose dimensionality is same as $\mathcal{M}$.
A Riemannian manifold is defined as a tuple $(\mathcal{M}, g)$ that is possessing metric tensor $g_x : \mathcal{T}_x\mathcal{M} \times \mathcal{T}_x\mathcal{M} \rightarrow \mathbb{R}$ on the tangent space $\mathcal{T}_x\mathcal{M}$ at each point $x \in \mathcal{M}$ \cite{petersen2006riemannian}.
The metric tensor provides geometric notions such as geodesic, angle and volume.
There exist mapping between the manifold and the tangent space: exponential map and logarithmic map.
The exponential map $\exp_x: \mathcal{T}_x\mathcal{M} \rightarrow \mathcal{M}$ projects the vector on the tangent space $\mathcal{T}_x\mathcal{M}$ back to the manifold $\mathcal{M}$, while the logarithmic map $\log_x: \mathcal{M} \rightarrow \mathcal{T}_x\mathcal{M}$ is the inverse mapping of the exponential map as  $\log_x(\exp_x(v))=v$. 

The hyperbolic space is a Riemannian manifold with constant negative sectional curvature equipped with hyperbolic geometry.
This paper deals with two hyperbolic spaces; \textit{`Poincar\'e ball'} and \textit{`hyperboloid'}.
The \textit{Poincar\'e ball} $\mathbb{P}$ is highly effective for visualizing and analyzing the hyperbolic latent space.
Meanwhile, the \textit{hyperboloid} $\mathbb{H}$ can provide stable optimization since, unlike distance function of \textit{Poincar\'e ball}, there is no division in the distance function \cite{nickel2018learning}. 
A review of Riemannian geometry and details of hyperboloid model are presented in the supplementary material.\\
\textbf{Poincar\'e ball.}
The $n$-dimensional Poincar\'e ball with constant negative curvature $K(K<0)$ $(\mathbb{P}_K^n, g_x^{\mathbb{P}_K})$ is defined:
\begin{equation}
\mathbb{P}_K^n = \{x \in \mathbb{R}^n : \|x\|^2 < -1/K \},
\end{equation} 
where $\| \cdot \|$ denotes Euclidean norm.
\begin{figure*}[t!]
\includegraphics[trim={3cm 38cm 3cm 2cm}, clip, width=.97\textwidth]{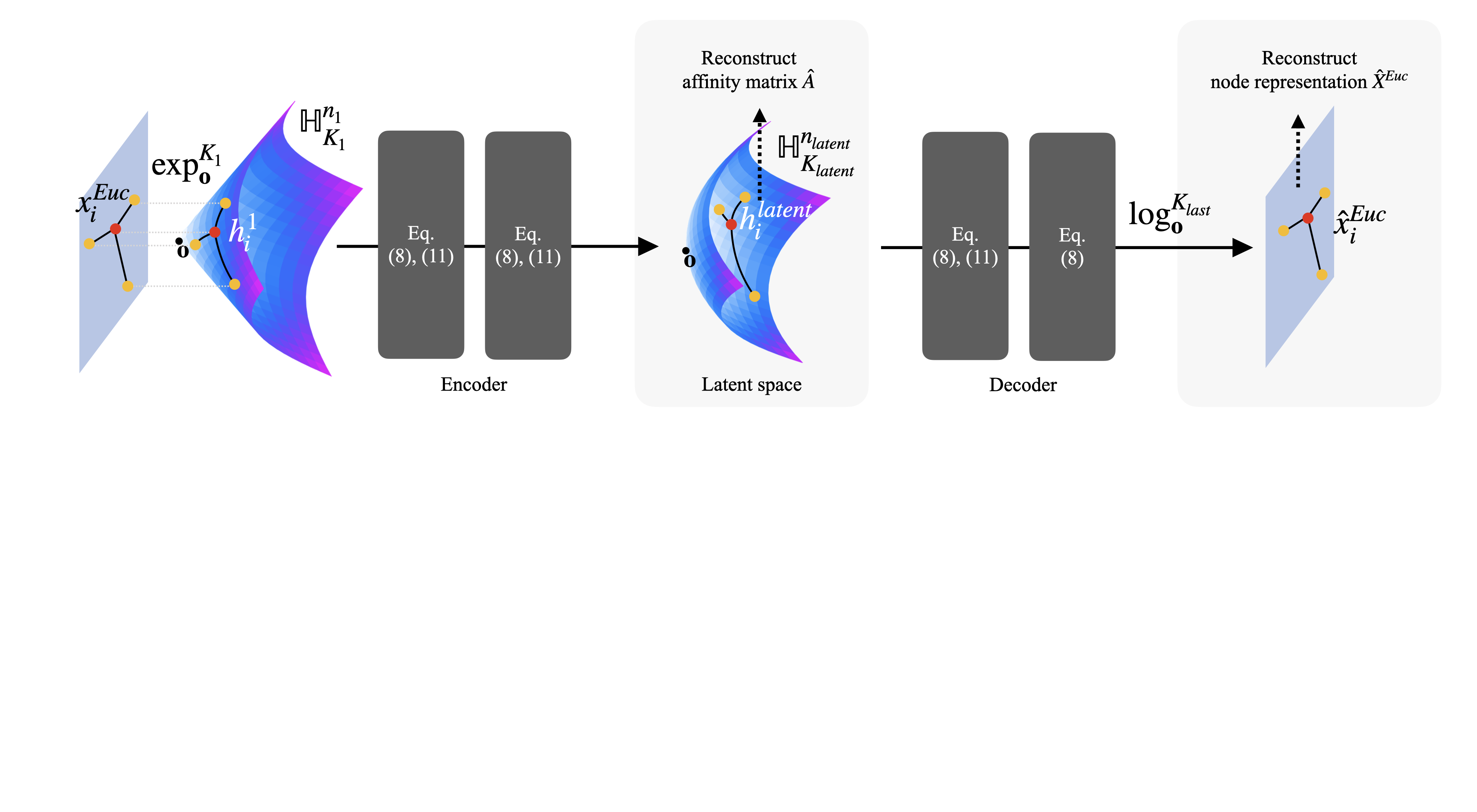}
\vspace{-2.5mm}
\caption{The overall architecture of HGCAE in a two-layer auto-encoder (\textit{i.e.} the encoder and decoder have two layers each) whose hyperbolic space is hyperboloid.
This figure describes three things: 1) how the node of the graph (red dot) conducts message passing (Eq. (\ref{messagepassing}) and (\ref{nonhyperboloid})) with its neighbors (yellow dot), 2) the process of embedding the output of encoder in hyperboloid latent space (blue-purple space), and 3) reconstruction of Euclidean node attributes at the end of the decoder.
}
\label{archi_hgae}
\vspace{-5mm}
\end{figure*}
The metric tensor is $g_x^{\mathbb{P}_K} = (\lambda_x^K)^2 g_x^\mathbb{E}$, where $\lambda_x^K = \frac{2}{1 + K\|x\|^2}$ is the conformal factor and $g_x^{\mathbb{E}} = \operatorname{diag}([1, 1, \ldots 1])$ denotes Euclidean metric tensor.
The origin of $\mathbb{P}_K^n$ is $\mathbf{o} = (0, \ldots , 0) \in \mathbb{R}^{n}$.
The distance between two points $x, y \in \mathbb{P}_K^n$ is defined as
{\small
\begin{equation}
d_{\mathbb{P}_K^n}(x,y) = \frac{1}{\sqrt{-K}}\operatorname{arcosh}\left(1- \frac{2K\|x-y\|^2}{(1+K\|x\|^2)(1+K\|y\|^2)}\right).
\label{dist_poincare}
\end{equation}}
For points $x \in \mathbb{P}_K^n$, tangent vector $v \in \mathcal{T}_x \mathbb{P}_K^n$, and $y \neq \textbf{0}$, the exponential map $\exp_x : \mathcal{T}_x \mathbb{P}_K^n \rightarrow \mathbb{P}_K^n$ and the logarithmic map $\log_x : \mathbb{P}_K^n \rightarrow \mathcal{T}_x \mathbb{P}_K^n$ are defined as:
{\small
\begin{align}
& \exp_x^K(v) = x \oplus_K \left(\tanh (\frac{\sqrt{-K}\lambda_x^K \|v\|}{2})\frac{v}{\sqrt{-K}\|v\|}\right), \\
& \log_x^K(y) = \frac{2}{\sqrt{-K}\lambda_x^K} \operatorname{arctanh}\left(\sqrt{-K}\|u\| \right) \frac{u}{\|u\|},
\end{align}}
where $u = -x \oplus_K y$ and $\oplus_K$ denotes M\"obius addition \cite{ungar2008gyrovector} for $x, y \in \mathbb{P}_K^n$ as
{\small
\begin{equation}
x \oplus_K y = \frac{(1-2K \langle x,y \rangle -K\| y\|^2)x + (1 +K\| x\|^2)y}{1-2K \langle x,y \rangle +K^2\|x\|^2\|y\|^2}.
\end{equation}}
\vspace{-3mm}\\

\noindent \textbf{Mapping between two models.} 
Two hyperbolic models, Poincar\'e ball and hyperboloid, are equivalent and transformations between two models retain many geometric properties including isometry.
There exist diffeomorphisms $p_{\mathbb{H}\rightarrow \mathbb{P}}$ and $p_{\mathbb{P} \rightarrow \mathbb{H}}$ between the two models, Poincar\'e ball $\mathbb{P}_K^n$ and hyperboloid $\mathbb{H}_K^n$ \cite{liu2019hyperbolic, chami2019hyperbolic}, as follows:
{\small
\begin{align}
& p_{\mathbb{H}\rightarrow \mathbb{P}}(x_0, x_1, \ldots , x_n) = \frac{(x_1, \ldots, x_n)}{\sqrt{|K|}x_0 + 1}, \label{H2P} \\
& p_{\mathbb{P} \rightarrow \mathbb{H}}(x_1, \ldots , x_n) = \frac{(\frac{1}{\sqrt{|K|}}(1-K\|x\|^2), 2x_1, \ldots, 2x_n)}{1+K\|x\|^2}. \label{P2H} 
\end{align}}
\vspace{-7mm}

\section{Methodology}
HGCAE is designed to fully utilize hyperbolic geometry in the auto-encoding process along with leveraging the power of graph convolutions via geometry-aware attention mechanism.
Each layer conducts message passing in hyperbolic space whose curvature value is trainable.
Before conducting message passing, we need to map the given input data points, $x^{Euc}$, defined in Euclidean space to the hyperbolic manifold.
We map the Euclidean feature into hyperbolic manifold via $h_i^1=\exp_{\textbf{o}}^{K_1}(x_i^{Euc})$, where $K_1$ and $h_i^1$ denote a trainable curvature value and the $i$-th node's representation of the first layer respectively.
When the hyperbolic space is hyperboloid model, we use $(0, x^{Euc}) \in \mathbb{R}^{n+1}$ as an input of an exponential map as \cite{chami2019hyperbolic} did.
The overall architecture of HGCAE is presented in Fig. \ref{archi_hgae}.

\subsection{Geometry-Aware Message Passing}
\noindent \textbf{Linear transformation.} Message passing in the HGCAE consists of two steps: the linear transformation of a message and aggregating messages from neighbors.
The $i$-th node's message passing result at the $l$-th layer $z_i^l$ is as follows:
\begin{equation}
z_i^{l} = \exp_{\textbf{o}}^{K_l} \left(\sum_{j \in \mathcal{N}(i)} \alpha_{ij}^l \Big(W^l \log_{\textbf{o}}^{K_l}(h_j^{l}) + b^l \Big) \right),
\label{messagepassing}
\end{equation}
where $W^l$, $b^l$, $\mathcal{N}(i)$, and $\alpha_{ij}^l$ denote a weight matrix, a bias term, the set of direct neighbors of node $i$ including itself, and the relative importance (attention score) of the neighbor node $j$ to the node $i$ at the $l$-th layer respectively.
Based on \cite{ganea2018hyperbolic}, we map the points in the hyperbolic manifold to the tangent space via the logarithmic map, since the linear transformation cannot be performed directly in hyperbolic spaces.
Then, the messages are linearly transformed on the tangent space of the origin in which inherits many properties of the ambient Euclidean space.

\noindent \textbf{Aggregation.} After performing linear transformation, 
we aggregate messages from neighbors via an attention mechanism.
The majority of message passing algorithms which use attention mechanisms learn the relative importance of each node's neighbors based on node feature not only in Euclidean space \cite{velickovic2018graph} but also in hyperbolic space \cite{chami2019hyperbolic}.
However, only considering node features for learning their relative importance does not take into account the geometry of the space, and this might result in an imprecise attention score.
To make full use of the Riemannian metric of the hyperbolic manifolds, we adopt a geometry-aware attention mechanism \cite{gulcehre2018hyperbolic} by utilizing the distance between linearly transformed node features on the hyperbolic space.
Let $y_i^l = W^l \log_{\textbf{o}}^{K_l}(h_i^{l}) + b^l$, then the attention score at the $l$-th layer in Eq. (\ref{messagepassing}) is:
\begin{equation}
\alpha_{ij}^l = \frac{\exp (-\beta^l d^2_{{\mathcal{M}_{K_l}}}(y_i^l, y_j^l)- \gamma^l)}{\sum_{p \in \mathcal{N}(i)}\exp (-\beta^l d^2_{{\mathcal{M}_{K_l}}}(y_i^l, y_p^l)-\gamma^l)},
\label{attention}
\end{equation}
where $d_{{\mathcal{M}_{K_l}}}(\cdot, \cdot)$, $\beta^l$, and $\gamma^l$ denote the distance on the hyperbolic space with curvature value $K_l$, and trainable parameters of the $l$-th layer respectively.
After every step of message passing, we map the representation on the tangent space to the hyperbolic manifold via the exponential map.

\subsection{Nonlinear Activation}
The nonlinear activations, $\sigma$, such as ReLU can be directly applied to the points in the Poincar\'e ball, in contrast to the points on the hyperboloid \cite{liu2019hyperbolic}.
Thus, when the hyperboloid model is used, we map the points to the Poincar\'e ball using Eq. (\ref{H2P}) first.
Next, we apply the nonlinear activation in the Poincar\'e ball, and then return the result to the hyperboloid using the Eq. (\ref{P2H}).

Since the curvature value of each layer in HGCAE is trainable, each layer can have different curvature values from other layers.
Thus, a step for locating the result of the nonlinear activation in the hyperbolic space having a curvature value of the next layer is required.
First, we map the results of the nonlinear activation to the tangent space of the current layer, $\mathcal{T}_{\mathbf{o}}\mathcal{M}_{K_l}$, using logarithmic map, $\log_{\mathbf{o}}^{K_l}$. 
Next, the points in the tangent space are mapped to the next layer's hyperbolic space via an exponential map of the next layer $\exp_{\mathbf{o}}^{K_{l+1}}$.
The equations for performing such nonlinear activation and mapping to the hyperbolic space of the next layer in the cases of Poincar\'e ball and hyperboloid are as follows respectively:
\begin{small}
\begin{align}
& h_i^{l+1} =  \exp_{\textbf{o}}^{K_{l+1}}\Big(\log_{\textbf{o}}^{K_l}\big(\sigma(z_i^{l})\big)\Big), \label{nonpoincare} \\
& h_i^{l+1} = \exp_{\textbf{o}}^{K_{l+1}}\Big(\log_{\textbf{o}}^{K_l}\big(p_{\mathbb{P} \rightarrow \mathbb{H}} ( \sigma ( p_{\mathbb{H} \rightarrow \mathbb{P}} (z_i^l)))\big)\Big). \label{nonhyperboloid}
\end{align}
\end{small}
\vspace{-5mm}

\subsection{Loss Function}
Our HGCAE reconstructs both the affinity matrix (graph structure) $A$ and the Euclidean node attributes $X^{Euc}$, at the end of the encoder and the decoder respectively.
To reconstruct the Euclidean node attributes $\hat{X}^{Euc}$, the aggregated representations in the hyperbolic space of the decoder's last layer are mapped to the tangent space of the origin $\mathcal{T}_{\mathbf{o}}\mathcal{M}$.
Then, the loss of representations $\mathcal{L}_{REC-X}$ is defined as the mean square error between $X^{Euc}$ and $\hat{X}^{Euc}$: $\frac{1}{N} \| X^{Euc} - \hat{X}^{Euc} \|^2$.
For reconstructing the structure of the graph, the hyperbolic distance between the latent representations (the output of the encoder) of two nodes is utilized.
To calculate the probability score of an edge which links between two nodes, we adopt the Fermi-Dirac distribution \cite{krioukov2010hyperbolic, nickel2017poincare}, $\hat{A}_{ij} = [e^{(d_{{\mathcal{M}_K}}^2(h_i , h_j)-r)/t}+1]^{-1}$, where $h_i$, $\hat{A}$, $r$, and $t$ denote the latent representation of node $i$, the reconstructed affinity matrix, and hyperparameters respectively.
The loss function for the affinity matrix is defined by the cross entropy loss with negative sampling: $\mathcal{L}_{REC-A} = \mathbb{E}_{q(H|X, A)}[\log p(\hat{A}|H)]$, where $q(H|X, A) = \prod_{i=1}^{N} q(h_i |X, A)$.
The overall loss function of HGCAE is
\begin{equation}
\mathcal{L} = \mathcal{L}_{REC-A} + \lambda \mathcal{L}_{REC-X},
\label{final loss}
\end{equation}
where $\lambda$ is a regularization parameter.
$\lambda$ serves to control the relative importance between the attributes and structure.

\begin{table}[t]
\caption{Dataset statistics.}
\vspace{-7mm}
\label{dataset}
\begin{center}
\footnotesize
\begin{tabularx}{0.45\textwidth}{lYYYY}
\midrule
Dataset             & Node   & Edge    & Attribute & Class \\
\midrule
Phylogenetic  \cite{hofbauer2016preliminary, sanderson1994treebase}         & 344    & 343     & -         & -  \\
CS PhDs \cite{de2018exploratory}             & 1,025  & 1,043   & -         & -  \\
Diseases \cite{goh2007human, nr-aaai15}            & 516    & 1,188   & -         & -  \\
Cora \cite{sen2008collective}                 & 2,708  & 5,429   & 1,433     & 7  \\
Citeseer \cite{sen2008collective}            & 3,312  & 4,552   & 3,703     & 6  \\
Wiki \cite{yang2015network}                & 2,405  & 17,981  & 4,973     & 17 \\
Pubmed \cite{sen2008collective}              & 19,717 & 44,338  & 500       & 3  \\
BlogCatalog \cite{tang2009relational}         & 5,196  & 171,743 & 8,189     & 6  \\
Amazon Photo \cite{mcauley2015image}        & 7,650  & 119,081 & 745       & 8  \\
ImageNet-10 \cite{chang2017deep}         & 13,000  &  -     & 27,648    & 10  \\
ImageNet-Dogs  \cite{chang2017deep}      & 19,500  &  -     & 27,648    & 15  \\
ImageNet-BNCR    & 11,700  &  -     & 27,648    & 9  \\
\midrule
\end{tabularx}
\label{dataset}
\end{center}
\vspace{-10mm}
\end{table}

\begin{table*}[t]
\caption{Link prediction performances.}
\vspace{-7mm}
\label{sample-table}
\begin{center}
\scriptsize
\begin{tabularx}{\textwidth}{lYYYYYYYYYYYY}
\midrule
\multirow{2}{*}{} & \multicolumn{2}{c}{Cora} & \multicolumn{2}{c}{Citeseer} & \multicolumn{2}{c}{Wiki} & \multicolumn{2}{c}{Pubmed} & \multicolumn{2}{c}{BlogCatalog} & \multicolumn{2}{c}{Amazon Photo} \\
\cmidrule{2-13}
 & AUC & AP & AUC & AP & AUC & AP & AUC & AP & AUC & AP & AUC & AP \\
\midrule        
GAE \cite{kipf2016variational}               & 0.910 & 0.920 & 0.895 & 0.899 & 0.930 & 0.948 & 0.964 & 0.965 & 0.840 & 0.841 & 0.956 & 0.948 \\
VGAE \cite{kipf2016variational}           & 0.914 & 0.926 & 0.908 & 0.920 & 0.936 & 0.950 & 0.944 & 0.947 & 0.844 & 0.846 & 0.971 & 0.966 \\
ARGA \cite{pan2018adversarially}              & 0.924 & 0.932 & 0.919 & 0.930 & 0.934 & 0.947 & \textbf{0.968} & 0.971 & 0.857 & 0.850 & 0.961 & 0.954 \\
ARVGA \cite{pan2018adversarially}             & 0.924 & 0.926 & 0.924 & 0.930 & 0.947 & 0.948 & 0.965 & 0.968 & 0.837 & 0.828 & 0.927 & 0.909 \\
GALA \cite{park2019symmetric}              & 0.929 & 0.937 & 0.944 & 0.948 & 0.936 & 0.931 & 0.915 & 0.897 & 0.774 & 0.765 & 0.918 & 0.910 \\
DBGAN \cite{zheng2020distribution}             & 0.945 & 0.951 & 0.945 & 0.958 & - & - & \textbf{0.968} & \textbf{0.973} & - & - & - & - \\
\midrule
\textbf{HGCAE-P}   & 0.948 & 0.947 & 0.960 & 0.963 & \textbf{0.955} & \textbf{0.962} & 0.962 & 0.960 & \textbf{0.896} & \textbf{0.886} & \textbf{0.982} & \textbf{0.976} \\
\textbf{HGCAE-H}   & \textbf{0.956} & \textbf{0.955} & \textbf{0.967} & \textbf{0.970} & 0.952 & 0.958 & 0.962 & 0.960 & 0.857 & 0.850 & 0.972 & 0.966 \\
\midrule
\end{tabularx}
\label{lp result}
\end{center}
\vspace{-6mm}
\end{table*}

\begin{table*}[t]
\caption{Node clustering performances.}
\vspace{-7mm}
\label{sample-table}
\begin{center}
\scriptsize
\begin{tabularx}{\textwidth}{lYYYYYYYYYYYY}
\midrule
\multirow{2}{*}{} & \multicolumn{2}{c}{Cora} & \multicolumn{2}{c}{Citeseer} & \multicolumn{2}{c}{Wiki} & \multicolumn{2}{c}{Pubmed} & \multicolumn{2}{c}{BlogCatalog} & \multicolumn{2}{c}{Amazon Photo} \\
\cmidrule{2-13}
 & ACC & NMI & ACC & NMI & ACC & NMI & ACC & NMI & ACC & NMI & ACC & NMI \\
\midrule
Kmeans \cite{lloyd1982least}               & 0.492 & 0.321 & 0.540 & 0.305 & 0.417 & 0.440 & 0.595 & 0.315 & 0.180 & 0.007 & 0.267 & 0.122  \\
GAE \cite{kipf2016variational}               & 0.532 & 0.434 & 0.505 & 0.246 & 0.460 & 0.468 & 0.686 & 0.295 & 0.284 & 0.112 & 0.390 & 0.337 \\
VGAE \cite{kipf2016variational}              & 0.595 & 0.446 & 0.467 & 0.260 & 0.450 & 0.467 & 0.688 & 0.310 & 0.269 & 0.097 & 0.418 & 0.376 \\
MGAE  \cite{wang2017mgae}             & 0.684 & 0.511 & 0.660 & 0.412 & 0.514 & 0.485 & 0.593 & 0.282 & 0.423 & 0.202 & 0.594 & 0.475 \\
ARGA  \cite{pan2018adversarially}             & 0.640 & 0.449 & 0.573 & 0.350 & 0.458 & 0.437 & 0.680 & 0.275 & 0.464 & 0.270 & 0.577 & 0.499 \\
ARVGA  \cite{pan2018adversarially}    & 0.638 & 0.450 & 0.544 & 0.261 & 0.386 & 0.338 & 0.513 & 0.116 & 0.450 & 0.250 & 0.455 & 0.395 \\
GALA   \cite{park2019symmetric}       & 0.745 & 0.576 & 0.693 & 0.441 & \textbf{0.544} & \textbf{0.503} & 0.693 & 0.327 & 0.400 & 0.251 & 0.512 & 0.485 \\
DBGAN  \cite{zheng2020distribution}   & 0.748 & 0.560 & 0.670 & 0.407 & - & - & 0.694 & 0.324 & - & - & - & - \\
\midrule
\textbf{HGCAE-P}                         & 0.746 & 0.572 & 0.693 & 0.422 & 0.459 & 0.467 & \textbf{0.748} & \textbf{0.377} & 0.550 & 0.325 & 0.781 & 0.696 \\
\textbf{HGCAE-H}                         & \textbf{0.767} & \textbf{0.599} & \textbf{0.715} & \textbf{0.453} & 0.530 & 0.435 & 0.711 & 0.347 & \textbf{0.741} & \textbf{0.578} & \textbf{0.817} & \textbf{0.722} \\
\midrule
\end{tabularx}
\label{nc result}
\end{center}
\vspace{-8mm}
\end{table*}

\begin{table}[t]
\caption{Link prediction task compared with P-VAE.}
\vspace{-7mm}
\begin{center}
\scriptsize
\begin{tabularx}{0.45\textwidth}{lYYYYYY}
\midrule
\multirow{3}{*}{} & \multicolumn{2}{c}{Phylogenetic} & \multicolumn{2}{c}{CS PhDs} & \multicolumn{2}{c}{Diseases} \\
\cmidrule{2-7}
& AUC & AP & AUC & AP & AUC & AP \\
\midrule        
VGAE \cite{kipf2016variational}              & 0.542 & 0.540 & 0.565 & 0.564 & 0.898 & 0.918 \\
P-VAE \cite{mathieu2019poincare}             & 0.590 & 0.555 & 0.598 & 0.567 & 0.923 & \textbf{0.936} \\
\midrule
\textbf{HGCAE-P}               & \textbf{0.688} & \textbf{0.712} & \textbf{0.673} & \textbf{0.640} & \textbf{0.926} & 0.914 \\
\midrule
\end{tabularx}
\label{lp pvae}
\end{center}
\vspace{-10mm}
\end{table}

\section{Experiments}
In this section, we explore the effectiveness of unsupervised hyperbolic embeddings on various data domains via quantitative and qualitative analyses.
We use $9$ real-world complex network datasets and $3$ image datasets. 
The statistics of the datasets are summarized in Table \ref{dataset}.
The details of the datasets, the compared methods, and the experimental details are described in the supplementary material.
For node clustering and link prediction tasks on the $9$ network datasets, we evaluate HGCAE-P and HGCAE-H, which denote HGCAE models whose latent spaces are Poincar\'é ball and hyperboloid respectively.
For the tasks of image clustering and visual data analysis, we use HGCAE-P because Poincar\'e ball is a powerful tool for visualizing and analyzing properties of hyperbolic visual embeddings.

\subsection{Node Clustering and Link Prediction}
\noindent \textbf{Comparison to embeddings in Euclidean latent space.} 
We evaluated the usefulness of hyperbolic representations by the performances of downstream tasks on citation \cite{sen2008collective, yang2015network}, social \cite{tang2009relational}, and co-purchase \cite{mcauley2015image} networks.
We compared against the state-of-the-art unsupervised message passing models \cite{kipf2016variational, pan2018adversarially, wang2017mgae, park2019symmetric, zheng2020distribution} which mainly conduct in Euclidean space.
Similar to evaluation metrics used in \cite{park2019symmetric}, we used area under curve (AUC) and average precision (AP) to evaluate the performance of the link prediction task, while using accuracy (ACC) and normalized mutual information (NMI) for evaluating the node clustering task.

The results of link prediction and node clustering are presented in Tables \ref{lp result} and \ref{nc result} respectively.
From the results, we can see that our HGCAE, with the representations of hyperbolic latent spaces, outperforms the existing methods, which use Euclidean latent spaces.
Our superior results over their Euclidean counterparts support the fact that unsupervised learning with message passing benefit from the geometry of hyperbolic spaces.
Due to space constraints, further analysis of the ablation study on the proposed architecture and the effectiveness of low-dimensional hyperbolic latent space are reported in the supplementary material.

\noindent \textbf{Comparison to embeddings of hyperbolic graph auto-encoder.}
To validate the architecture of HGCAE, we compared its performance with the Poincar\'e variational auto-encoder (P-VAE) \cite{mathieu2019poincare}, whose latent space is the Poincar\'e ball and conducts its message passing in Euclidean space.
Three networks, phylogenetic tree \cite{hofbauer2016preliminary, sanderson1994treebase}, Ph.D. advisor-student relationships \cite{de2018exploratory}, and disease relationships \cite{goh2007human, nr-aaai15}, were used for evaluating performance on link prediction.
The latent space of both P-VAE and HGCAE-P is a $5$-dimensional Poincar\'e ball.
We report the results in Table \ref{lp pvae}.
The proposed HGCAE-P outperforms P-VAE for most cases of the datasets since HGCAE-P considers hyperbolic geometry in the whole auto-encoding processes.

\begin{figure}[t!]
\centering
\includegraphics[trim={1.3cm 51cm 11cm 0.5cm}, clip, width=.45\textwidth]{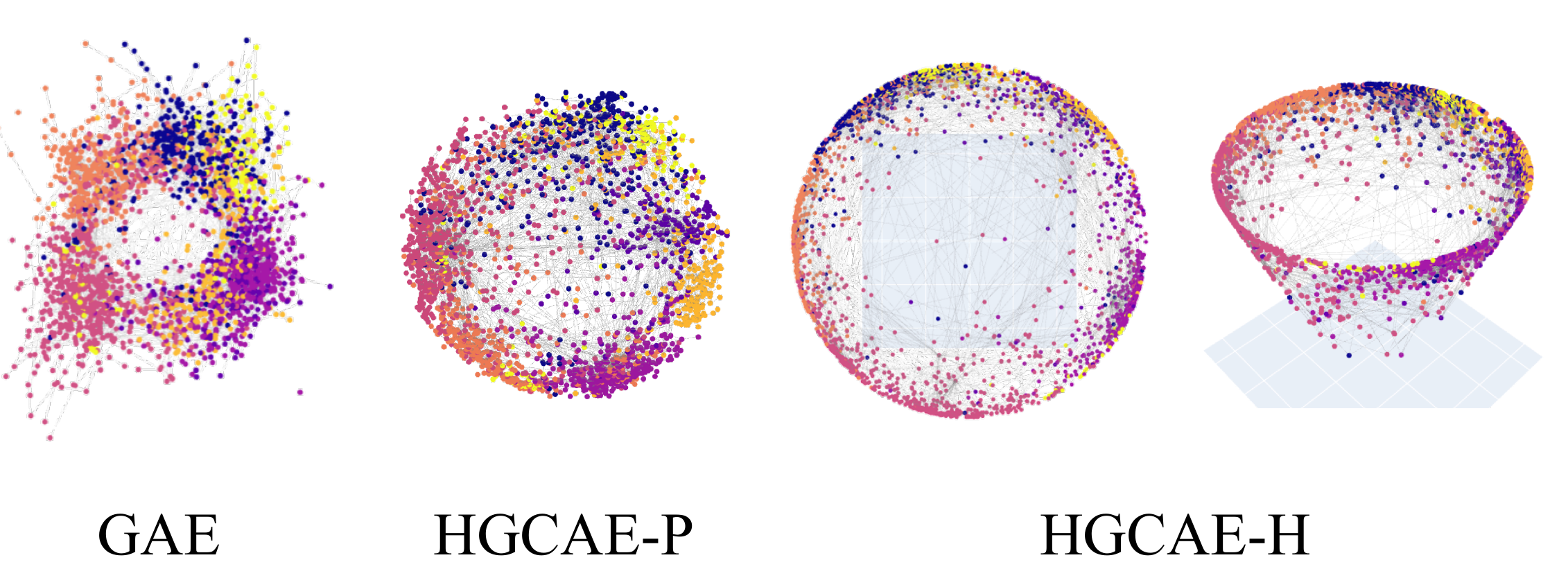}
\vspace{-2.5mm}
\caption{$2$-dimensional embeddings in Euclidean, Poincar\'e ball, and hyperboloid latent space on Cora dataset.
Same color indicates same class.
On hyperbolic latent spaces, most of the nodes are located on the boundary and well-clustered with the nodes in the same class.
}
\label{cora_embedding}
\vspace{-6mm}
\end{figure}
\noindent \textbf{Visualization of citation network.}
We explored the latent representations of GAE \cite{kipf2016variational} and our models on the Cora dataset \cite{sen2008collective} by constraining the latent space as a $2$-dimensional hyperbolic or Euclidean space.
The result is given in Fig. \ref{cora_embedding}.
On the results of HGCAE, most of the nodes are located on the boundary of hyperbolic spaces and well-clustered with the nodes in the same class.
Further visualization of the network datasets is presented in the supplementary material.

\subsection{Image Clustering}
\begin{figure*}[htb!]
\begin{minipage}[b]{.5\linewidth}
  \centering
  \centerline{\includegraphics[trim={0cm 0.45cm 1cm 0}, width=0.95\linewidth]{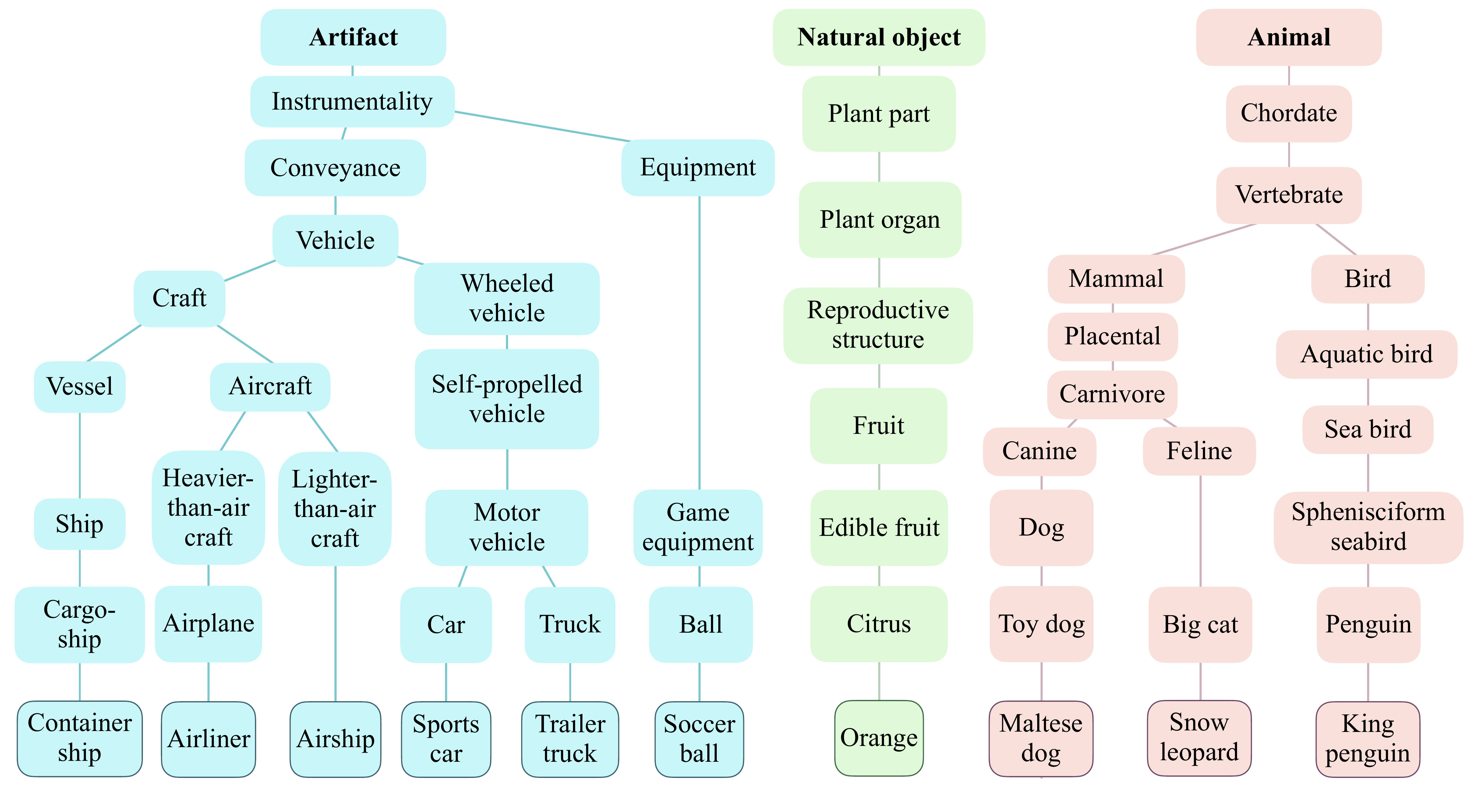}}
  \centerline{(a) ImageNet-10}\medskip
\end{minipage}
\hfill
\begin{minipage}[b]{0.5\linewidth}
  \centering
  \centerline{\includegraphics[trim={0cm 0.45cm 1cm 0}, width=0.95\linewidth]{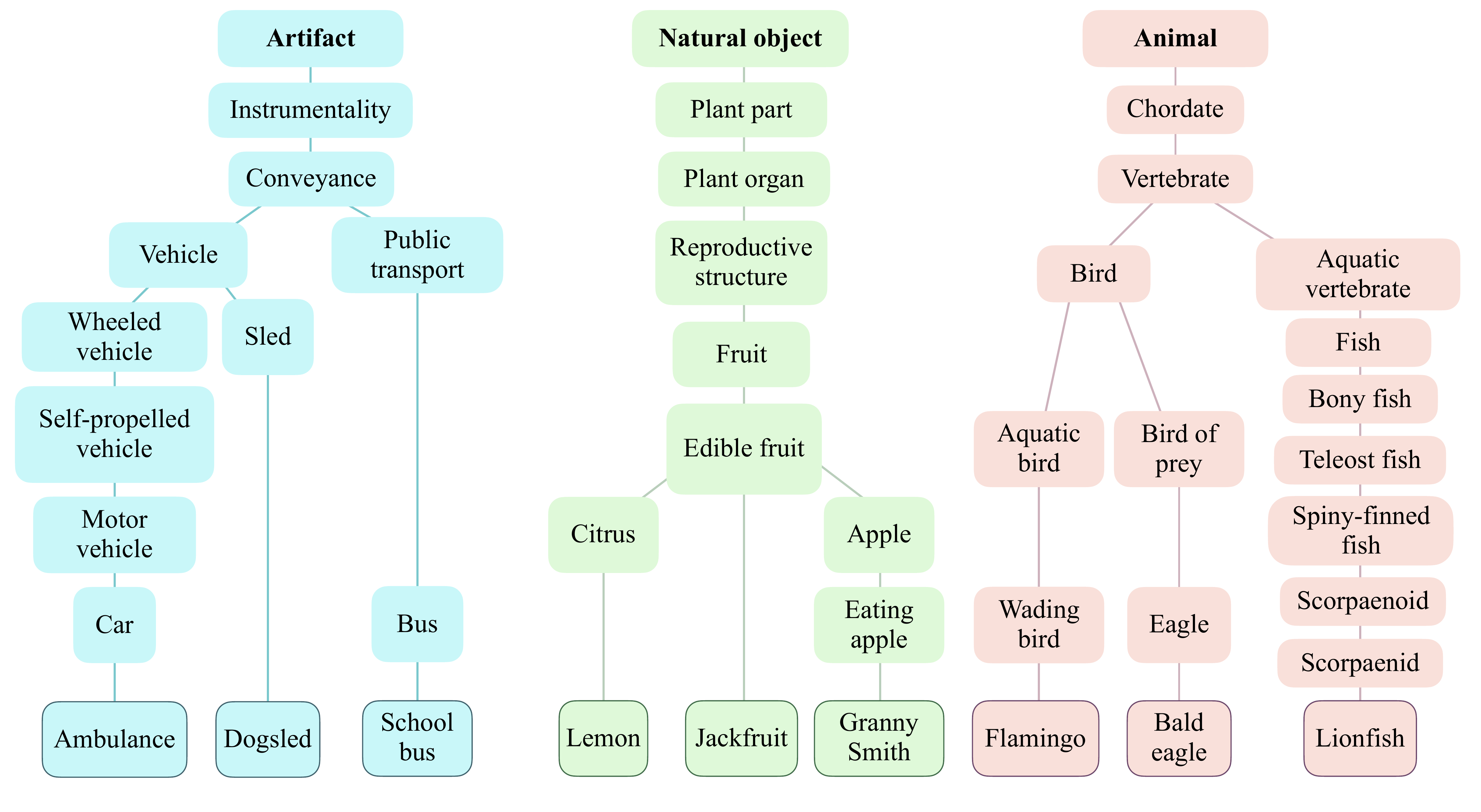}}
  \centerline{(b) ImageNet-BNCR}\medskip
\end{minipage}
\vspace{-9mm}
\caption{Class hierarchy of ImageNet-10 and ImageNet-BNCR\protect\footnotemark.
}
\label{tree class}
\vspace{-5mm}
\end{figure*}
\footnotetext{\url{http://image-net.org/index}}
In this experiment, we illustrate that image clustering can benefit from hyperbolic geometry. 
The training sets of ImageNet-10 and ImageNet-Dogs \cite{chang2017deep}, which are subsets of ImageNet \cite{krizhevsky2012imagenet}, are used for evaluation. 
In the manner of the researches \cite{ganea2018hyperbolic, gulcehre2018hyperbolic, khrulkov2020hyperbolic} which impose hyperbolic geometry on the activations of neural networks, we used the activations of PICA \cite{huang2020deep}, one of the most recent models developed for deep image clustering. 
After obtaining activations from the pre-trained networks of PICA, we built the graph by mutual $k$ nearest neighbors between activations. \begin{table}[t]
\caption{Image clustering performances.}
\vspace{-7mm}
\label{sample-table}
\begin{center}
\scriptsize
\begin{tabularx}{0.45\textwidth}{lYYYYYY}
\midrule
\multirow{2}{*}{} & \multicolumn{3}{c}{ImageNet-10} & \multicolumn{3}{c}{ImageNet-Dogs} \\
\cmidrule{2-7}
 & ACC & NMI & ARI & ACC & NMI & ARI \\
\midrule
Kmeans \cite{lloyd1982least}            & 0.241 & 0.119 & 0.057 & 0.105 & 0.055 & 0.020 \\
SC  \cite{zelnik2005self}               & 0.274 & 0.151 & 0.076 & 0.111 & 0.038 & 0.013 \\
AC  \cite{gowda1978agglomerative}       & 0.242 & 0.138 & 0.067 & 0.139 & 0.037 & 0.021 \\
NMF  \cite{cai2009locality}             & 0.230 & 0.132 & 0.065 & 0.118 & 0.044 & 0.016 \\
AE \cite{bengio2007greedy}              & 0.317 & 0.210 & 0.152 & 0.185 & 0.104 & 0.073 \\
CAE \cite{masci2011stacked}             & 0.253 & 0.134 & 0.068 & 0.134 & 0.059 & 0.022 \\
SAE  \cite{ng2011sparse}                & 0.325 & 0.212 & 0.174 & 0.183 & 0.112 & 0.072 \\
DAE  \cite{vincent2010stacked}          & 0.304 & 0.206 & 0.138 & 0.190 & 0.104 & 0.078 \\
DCGAN \cite{radford2015unsupervised}    & 0.346 & 0.225 & 0.157 & 0.174 & 0.121 & 0.078 \\
DeCNN \cite{zeiler2010deconvolutional}  & 0.313 & 0.186 & 0.142 & 0.175 & 0.098 & 0.073 \\
SWWAE \cite{zhao2015stacked}            & 0.323 & 0.176 & 0.160 & 0.158 & 0.093 & 0.076 \\
VAE  \cite{kingma2013auto}              & 0.334 & 0.193 & 0.168 & 0.179 & 0.107 & 0.079 \\
JULE \cite{yang2016joint}               & 0.300 & 0.175 & 0.138 & 0.138 & 0.054 & 0.028 \\
DEC \cite{xie2016unsupervised}          & 0.381 & 0.282 & 0.203 & 0.195 & 0.122 & 0.079 \\
DAC \cite{chang2017deep}                & 0.527 & 0.394 & 0.302 & 0.275 & 0.219 & 0.111 \\
DDC \cite{chang2019deep}                & 0.577 & 0.433 & 0.345 & -     & -     & -     \\
DCCM \cite{wu2019deep}                  & 0.710 & 0.608 & 0.555 & 0.383 & 0.321 & 0.182 \\
PICA$^{\dagger}$\cite{huang2020deep}    & 0.850 & 0.782 & 0.733 & 0.324 & 0.336 & 0.179 \\
PICA$^{\ddagger}$\cite{huang2020deep}   & 0.828 & 0.763 & 0.692 & 0.352 & 0.353 & 0.201 \\
\midrule
PICA$^{\ddagger}$\cite{huang2020deep}+HAE     & 0.821 & 0.759 & 0.686 & 0.338 & 0.347 & 0.200  \\
PICA$^{\ddagger}$\cite{huang2020deep}+GAE \cite{kipf2016variational}     & 0.854 & \textbf{0.792} & 0.737 & 0.344 & 0.350 & 0.199 \\
\midrule
\textbf{PICA$^{\ddagger}$\cite{huang2020deep}+HGCAE-P} & \textbf{0.855} & 0.790 & \textbf{0.741} & \textbf{0.387} & \textbf{0.360} & \textbf{0.226} \\
\midrule
\end{tabularx}
\end{center}
\vspace{-4mm}
\footnotesize{$^{\dagger}$ Numbers from literature. \\
$^{\ddagger}$ Numbers from our experiments on the official pre-trained networks\tablefootnote{\url{https://github.com/Raymond-sci/PICA}}.}
\label{ic result}
\vspace{-5mm}
\end{table}
Then, both the activations and the graph were used as inputs of HGCAE-P. 
Extensive baselines and state-of-the-art image clustering methods \cite{lloyd1982least, zelnik2005self, gowda1978agglomerative, cai2009locality, bengio2007greedy, masci2011stacked, ng2011sparse, vincent2010stacked, radford2015unsupervised, zeiler2010deconvolutional, zhao2015stacked, kingma2013auto, yang2016joint, xie2016unsupervised, chang2017deep, chang2019deep, wu2019deep, huang2020deep} were compared.
Furthermore, we also trained two auto-encoder models, GAE \cite{kipf2016variational}, and hyperbolic auto-encoder (HAE) whose layers are hyperbolic feed-forward layers \cite{ganea2018hyperbolic}.
The image clustering results are reported in Table \ref{ic result}.
The metrics, ACC, NMI and Adjusted Rand Index (ARI), were used for evaluation.
The results demonstrate that applying hyperbolic geometry along with using additional information of the approximated image manifold via nearest neighbor graphs can achieve better results than the Euclidean counterparts. 
We can also observe that HAE, the auto-encoder which naively applies hyperbolic geometry, does not work well, while our model performs better via the message passing fully utilizing hyperbolic geometry. 
\vspace{-1mm}

\subsection{Structure-Aware Unsupervised Embeddings}
In this experiment, we observe the unsupervised hyperbolic image embeddings' ability to recognize the latent structure of visual datasets that have hierarchical structures.
ImageNet \cite{krizhevsky2012imagenet} is constructed following the hierarchy of WordNet \cite{miller1998wordnet}, therefore, its classes of ImageNet-10 \cite{chang2017deep} also have hierarchical structures.
However, it is difficult to explore the effectiveness of hyperbolic embeddings since the classes of ImageNet-10 are biased to a certain root.
Thus, we have constructed a new dataset, \textit{ImageNet-BNCR}, that has a \textit{B}alanced \textit{N}umber of \textit{C}lasses across \textit{R}oots.
For \textit{ImageNet-BNCR}, we have chosen three roots, \textit{Artifact, Natural objects}, and \textit{Animal}, which have a large number of leaf classes.
Each root contains balanced child nodes of \textit{\{Ambulance, Dogsled, School bus\}}, \textit{\{Lemon, Jackfruit, Granny Smith\}}, and \textit{\{Flamingo, Bald eagle, Lionfish\}}, respectively.
On the leaf classes of ImageNet-10, \textit{\{Container ship, Airliner, Airship, Sports car, Trailer truck, Soccer ball\}}, \textit{\{Orange\}}, and \textit{\{Maltese dog, Snow leopard, King penguin\}} are the child nodes of the roots \textit{Artifact, Natural objects}, and \textit{Animal}, respectively. 
The class hierarchies of ImageNet-10 and ImageNet-BNCR are shown in Fig. \ref{tree class}.

\begin{figure}[t!]
\centering
\includegraphics[trim={0cm 15.7cm 12cm 0}, clip, width=.44\textwidth]{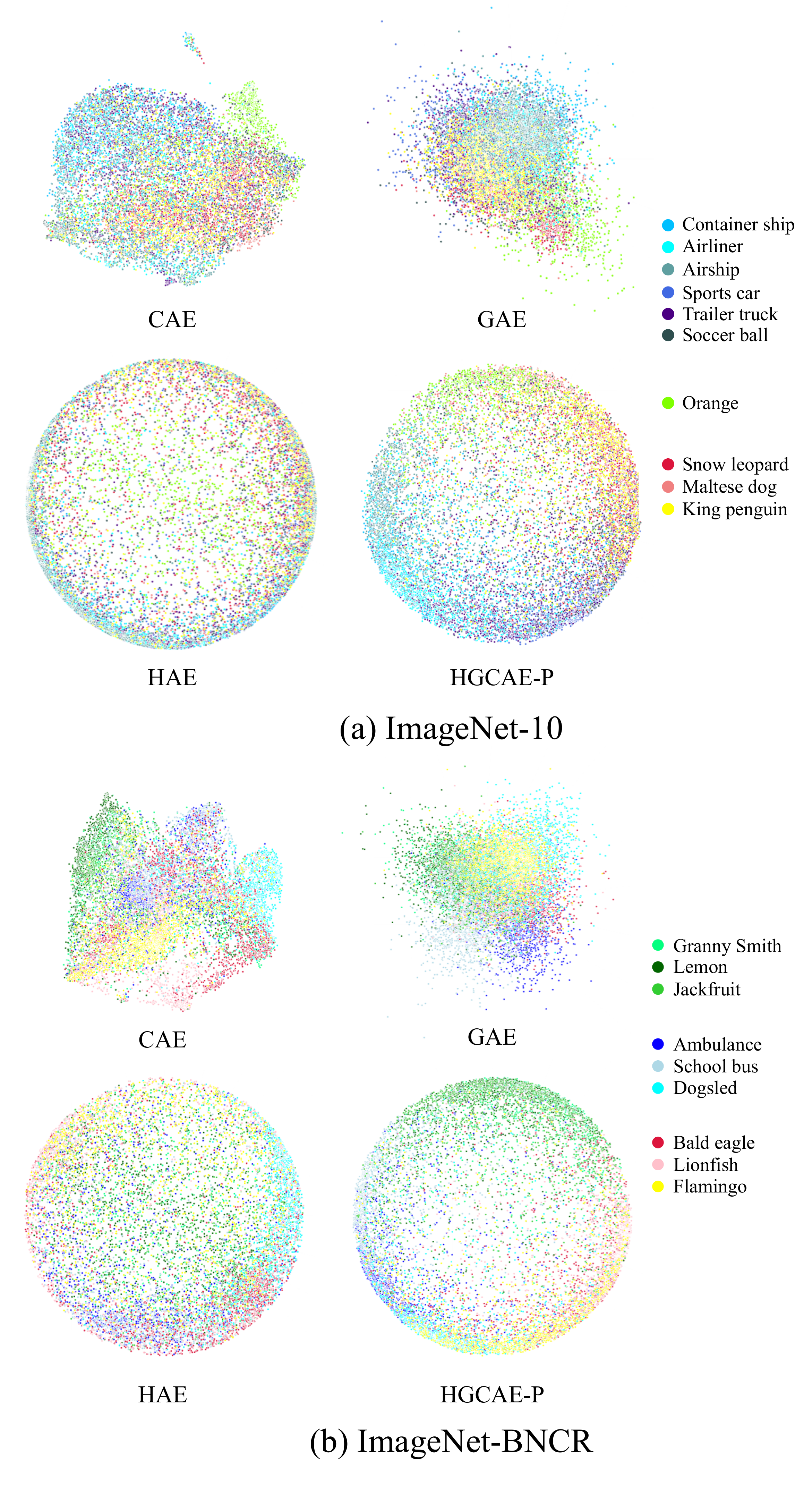}
\vspace{-7mm}
\caption{$2$-dimensional embeddings of CAE, GAE, HAE, and HGCAE-P on ImageNet-10 and ImageNet-BNCR.
Hyperbolic representations belonging to the same root are close to each other near the boundary of the space.}
\label{hierarchical_embedding}
\vspace{-7.5mm}
\end{figure}
\begin{figure}[t!]
\centering
\includegraphics[trim={1cm 41.7cm 14cm 0}, clip, width=.44\textwidth]{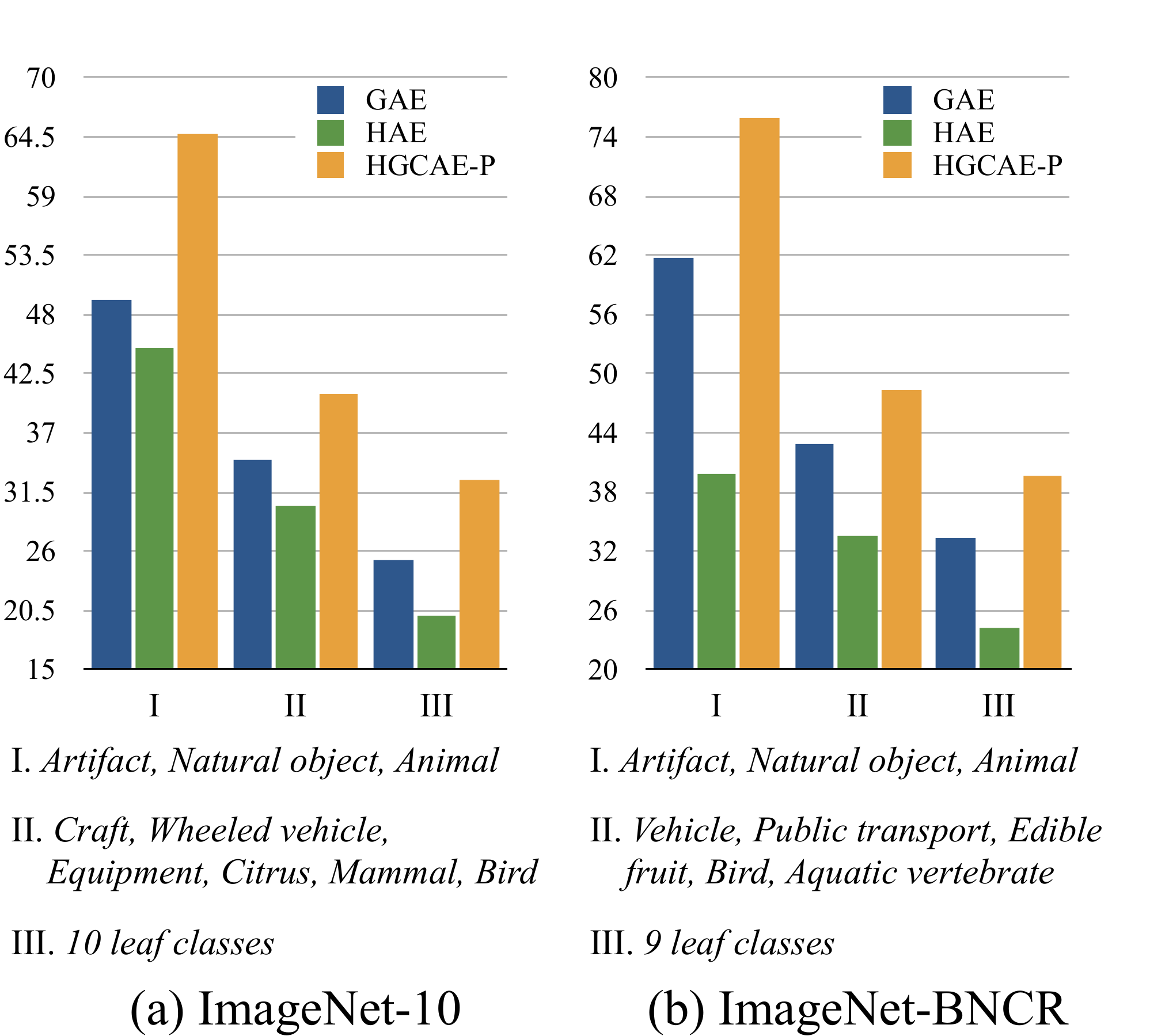}
\vspace{-2.5mm}
\caption{Clustering accuracy (\%) according to the hierarchy of classes on ImageNet-10 and ImageNet-BNCR.}
\label{hierarchy clustering}
\vspace{-6mm}
\end{figure}
\begin{figure}[t!]
\includegraphics[trim={0cm 12.9cm 33cm 0}, clip, width=.44\textwidth]{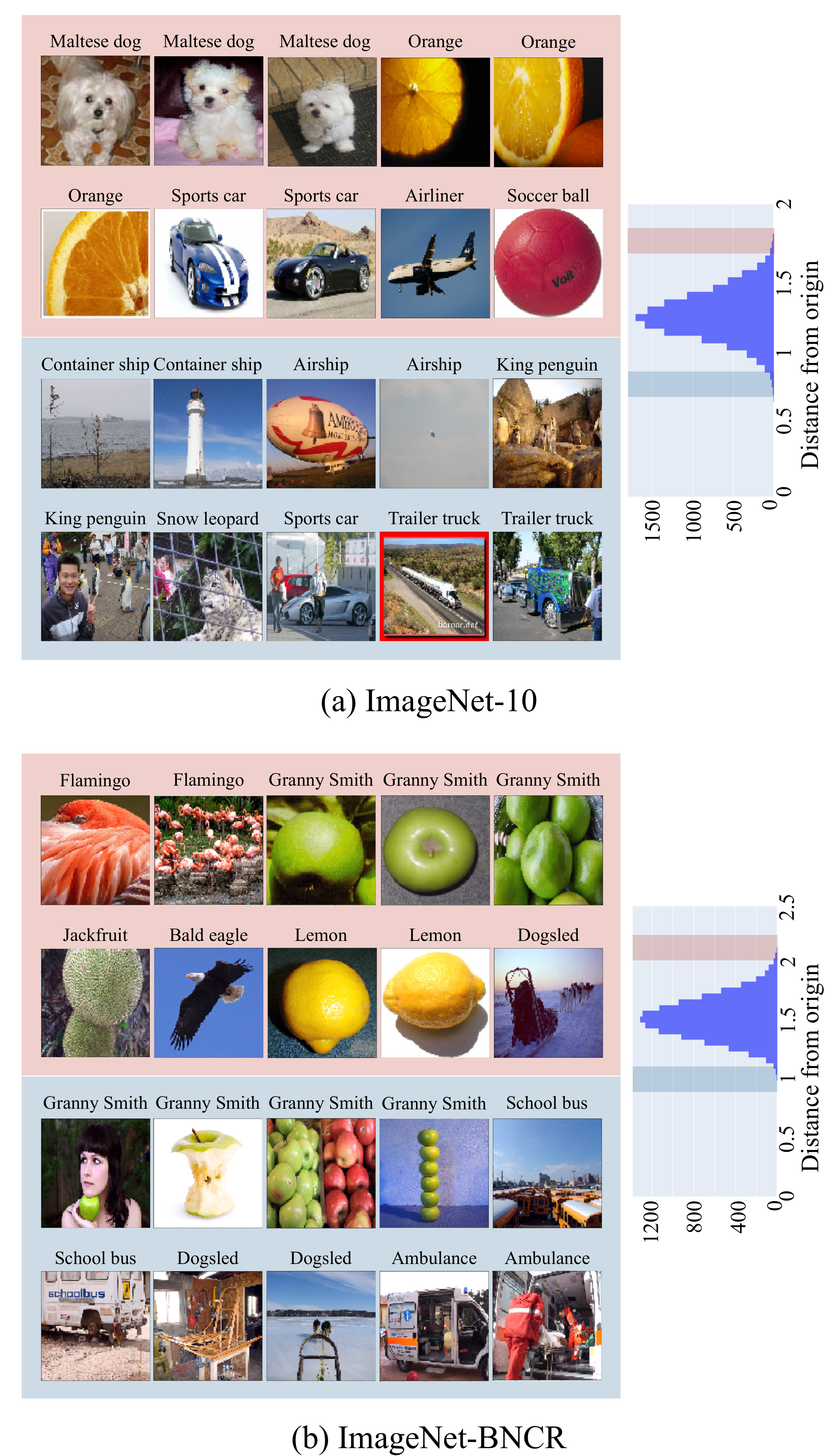}
\vspace{-2.5mm}
\caption{Histogram and images according to the hyperbolic distance from the origin (HDO) on ImageNet-10 and ImageNet-BNCR.
The feature of images inside red (blue) color box have high (low) HDO, so are located near the boundary (origin) of hyperbolic space.}
\label{confidence}
\vspace{-6mm}
\end{figure}
\begin{figure}[t!]
\centering
\includegraphics[trim={0.5cm 41cm 14cm 0}, clip, width=.44\textwidth]{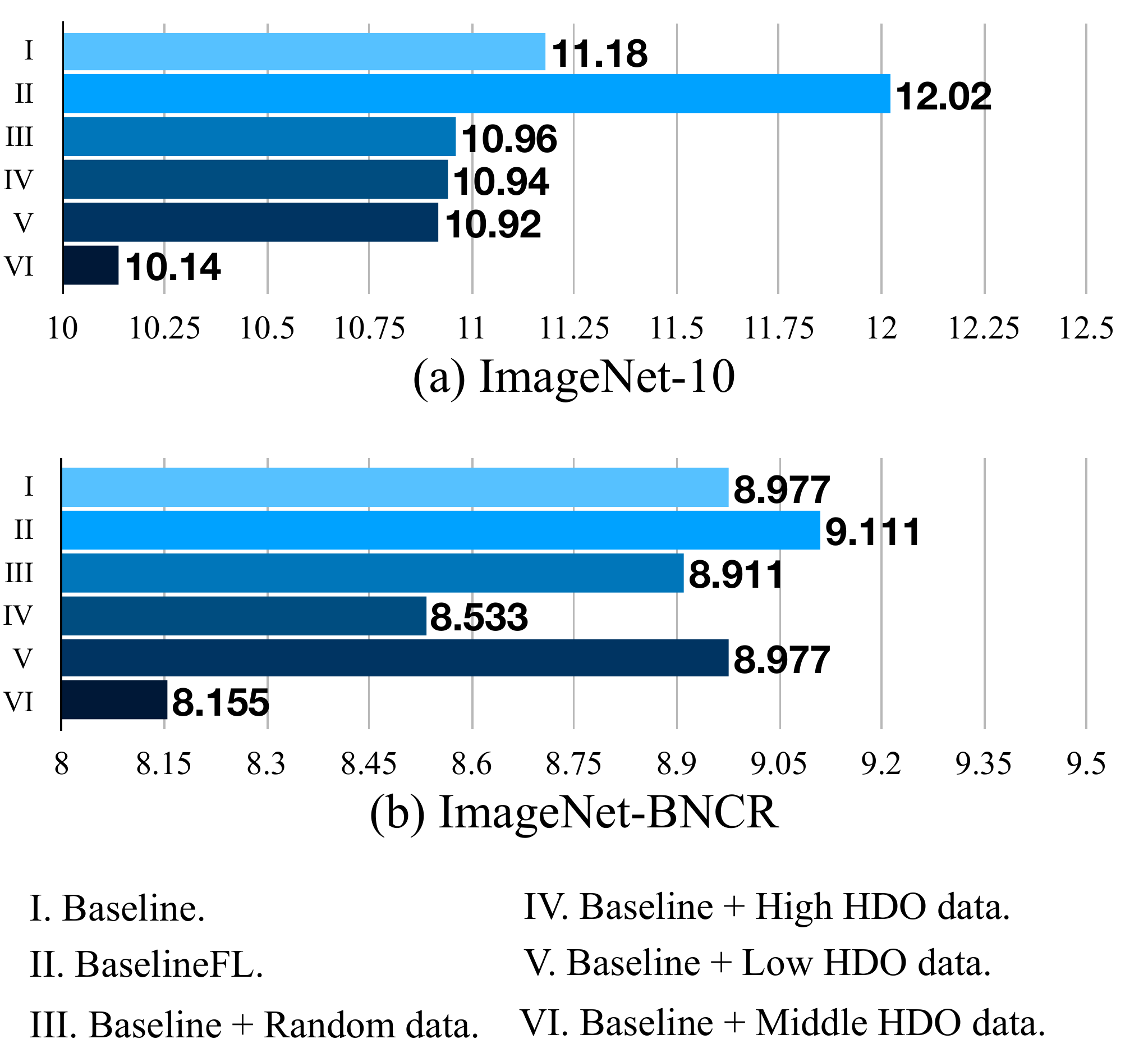}
\vspace{-2.5mm}
\caption{Top-1 classification error (\%) on ImageNet-10 and ImageNet-BNCR.}
\label{classification}
\vspace{-6.2mm}
\end{figure}
We extracted $1000$-dimensional features by training a convolutional auto-encoder (CAE) \cite{masci2011stacked} on the ImageNet-10 and ImageNet-BNCR datasets.
Then, after building the graph using mutual $k$ nearest neighbors between extracted features, we trained three auto-encoder models (HGCAE-P, GAE \cite{kipf2016variational}, and HAE) whose latent space is $2$-dimensional without the ground truth hierarchy structure of labels.
The embedding results of the $1000$-dimensional CAE features via UMAP \cite{mcinnes2018umap} and three auto-encoders are presented in Fig. \ref{hierarchical_embedding}.
We can observe that the embeddings of HGCAE-P are better clustered than others, according to the classes of each root in Fig. \ref{tree class}.
On the ImageNet-10, in the same root \textit{Artifact}, the embeddings of descendants of \textit{Craft} and \textit{Wheeled vehicle} are clustered respectively.
The embeddings of the ImageNet-BNCR are clustered more distinctly according to the root of class hierarchy than with ImageNet-10.
On the other hand, the embeddings of the root \textit{Natural objects}, \textit{\{Lemon, Jackfruit, Granny Smith\}}, are located closer to each other, since geodesic distance between each leaf label is small.
Our distinction from HAE implies that the additional information on image manifolds approximated by nearest neighbor graphs is helpful.
In contrast to the representations of CAE and GAE, we can see that the hyperbolic representations belonging to the same root are located near the boundary of the space.
In addition, to quantitatively validate the ability to recognize latent hierarchical structure of the data without direct learning of label hierarchy,
we cluster $2$-dimensional embeddings of the three auto-encoders with three ground truth label settings according to the class hierarchy in Fig. \ref{tree class}: I. Root nodes, II. Internal nodes, and III. Leaf nodes.
The quantitative results (clustering accuracy) on ImageNet-10 and ImageNet-BNCR are reported in Fig. \ref{hierarchy clustering}.
HGCAE-P outperforms GAE and HAE in every label hierarchy settings.
This might be because the leaf classes whose parent is the same are closely embedded with each other.
This analysis empirically demonstrates that unsupervised hyperbolic image embeddings can recognize the latent structure of the visual data that has a hierarchical structure.
\vspace{-1mm}

\subsection{Hyperbolic Distance to Filter Training Samples} 
In this experiment, we show that hyperbolic distance can help to choose training samples beneficial to the generalization ability of neural networks.
To this end, we obtained the latent embeddings of ImageNet-10 \cite{chang2017deep} and ImageNet-BNCR via HGCAE-P model.
Then, the hyperbolic distance (Eq. (\ref{dist_poincare})) of each embedding from the origin was computed.
Fig. \ref{confidence} shows some samples near the boundary or near the origin in the histogram of the hyperbolic distance from embeddings to the origin. 
We can see that the samples near the boundary can be easily classified, whereas those near the origin are harder to classify.
In general, the easy samples are not influential to learn an exact decision boundary.
On the other hand, the hard samples make the decision boundary over-fitted, i.e., they work like noises located at the soft margin region near the decision boundary \cite{cortes1995support}. 
This illustration intuitively shows that the \textit{H}yperbolic \textit{D}istance from the \textit{O}rigin (HDO) of a sample could give a clue which samples are influential or beneficial to learn the decision boundary crucial for the generalization ability of a classifier.

To verify this intuition, we conducted an experiment on the image classification task.
On ImageNet-10 and ImageNet-BNCR, we trained the VGG-11 \cite{simonyan2014very} classifier by adding further samples near the boundary/median of the distance histogram/origin to the original dataset in every training epoch and evaluated the network via each class' validation set in ImageNet \cite{krizhevsky2012imagenet}.
We compared our results with six settings: 
I. Baseline: original data with cross entropy loss, 
II. BaselineFL: original data with focal loss (FL) \cite{lin2017focal}\footnote{The focal loss tries to focus gradient updates on the samples that the classifier hard to classify.},
III. Baseline $+$ Random data adding, 
IV. Baseline $+$ High HDO data adding,
V. Baseline $+$ Low HDO data adding, and
VI. Baseline $+$ Middle HDO data adding.

The image classification results are given in Fig. \ref{classification}.
As expected, the case V of adding low HDO data in the histogram show similar performances with the baseline.
The case IV of adding high HDO data contributes the performance improvements, but the case VI of adding middle HDO data demonstrates the best performance among the compared settings. 
This result empirically verifies that the middle HDO samples are beneficial to learn a reasonable decision boundary which increases the generalization ability of a neural network.
Since the supporting samples marginally apart from the decision boundary are crucial for the generalization performance \cite{cortes1995support}, the HDO related with the generalization performance can be interpreted as a measure proportional to the distance of a sample from the decision boundary for a given classification task. 
In conclusion, we can utilize HDO as a criterion to choose samples for improving the generalization ability of a model for a given dataset.
\vspace{-2mm}

\section{Conclusion}
\vspace{-0.7mm}
In this paper, we explored the properties of unsupervised hyperbolic representations.
We derived the representations from geometry-aware message passing auto-encoders whose whole operations were conducted in hyperbolic spaces.
Then, we conducted extensive experiments and analyses on the low-dimensional latent representations in hyperbolic spaces.
The experimental results support the conclusion that taking advantage of hyperbolic geometry can improve the performances of unsupervised tasks; node clustering, link prediction, and image clustering.
We observed that the proposed method could yield unsupervised hyperbolic image embeddings reflecting the latent structure of the visual datasets that have hierarchical structure.
Lastly, we demonstrated that the hyperbolic distance from origin for a sample could be utilized to determine the additional data crucial for generalization ability of a classifier.

{\footnotesize
\noindent\textbf{Acknowledgement:} 
We thank Esha Dasgupta for careful reading and insightful comments on our manuscript.
This work was supported by Ministry of Science and ICT, Korea: IITP grant [No.2014-3-00123, Development of High Performance Visual BigData Discovery Platform] 
and ITRC support program [IITP-2020-2020-0-01789] supervised by the IITP.
}

{\small
\bibliographystyle{ieee_fullname}
\bibliography{egbib}
}

\clearpage
\twocolumn[
\centering
\Large
\textbf{Supplementary material for ``Unsupervised Hyperbolic Representation Learning via Message Passing Auto-Encoders"}
\vspace{1.0cm}
]

\appendix

In this supplemental material, we present the reviews of Riemannian geometry and hyperboloid model firstly.
Then, we explain the details of the datasets, compared methods, and experimental details.
Finally, further experiments on network datasets and further discussions are presented.

\section{Riemannian Geometry}
\subsection{A Review of Riemannian Geometry}
A manifold $\mathcal{M}$ of $n$-dimension is a topological space that each point $x \in \mathcal{M}$ has a neighborhood that is homeomorphic to $n$-dimensional Euclidean space $\mathbb{R}^n$.
For each point $x \in \mathcal{M}$, a real vector space $\mathcal{T}_x\mathcal{M}$ whose dimensionality is the same as $\mathcal{M}$ exists and is called a tangent space.
The tangent space $\mathcal{T}_x\mathcal{M}$ is the set of all the possible directions and speeds of the curves on $\mathcal{M}$ across $x \in \mathcal{M}$.
A Riemannian manifold is a tuple $(\mathcal{M}, g)$ that is possessing Riemannian metric $g_x : \mathcal{T}_x\mathcal{M} \times \mathcal{T}_x\mathcal{M} \rightarrow \mathbb{R}$ on the tangent space $\mathcal{T}_x\mathcal{M}$ at each point $x \in \mathcal{M}$ such that $\langle y, z \rangle_x = g_x(y, z) = y^TG(x)z$, where $G(x)$ is a matrix representation of Riemannian metric \cite{petersen2006riemannian}.
The metric tensor provides geometric notions such as the length of curve, angle and volume.
The length of curve $\gamma: t \mapsto \gamma(t) \in \mathcal{M}$ is $L(\gamma) = \int_{0}^{1} \| \gamma^{'}(t) \|_{\gamma(t)}^{1/2}  \,dt$.
The geodesic, the generalization of straight line on Euclidean space, is the constant speed curves giving the shortest path between the pair of points $x, y \in \mathcal{M}$: $\gamma^* = \operatorname*{arg\,min}_\gamma L(\gamma)$ where $\gamma(0) = x$, $\gamma(1) = y$ and $\| \gamma^{'}(t) \|_{\gamma(t)} = 1$.
The global distance between two points $x, y \in \mathcal{M}$ is defined as $d_{\mathcal{M}}(x, y) = \inf_{\gamma} L(\gamma)$.
For a tangent vector $v \in \mathcal{T}_x\mathcal{M}$ of $x \in \mathcal{M}$, there exists a unique unit speed geodesic $\gamma$ such that $\gamma(0) = x$ and $\gamma^{'}(0) = v$.
Then, the corresponding exponential map is defined as $\exp_x(v) = \gamma(1)$.
The inverse mapping of exponential map, the logarithmic map, is defined as $\log_x: \mathcal{M} \rightarrow \mathcal{T}_x\mathcal{M}$.
Refer the website of footnote for good introduction of hyperbolic geometry\footnote{\url{http://hyperbolicdeeplearning.com/simple-geometry-initiation/}}.

\subsection{Hyperboloid Model}
The hyperbolic space is a Riemannian manifold with constant negative sectional curvature equipped with hyperbolic geometry, and the hyperboloid model is one of the multiple equivalent hyperbolic models.
For $x, y \in \mathbb{R}^{n+1}$, the Lorentz inner product $\langle \cdot, \cdot \rangle_\mathcal{L}$ is defined as $\langle x, y \rangle_\mathcal{L} = -x_0 y_0 + \sum_{i=1}^n x_i y_i$.
The $n$-dimensional hyperboloid with constant negative curvature $K (K < 0)$ is defined as $(\mathbb{H}_K^n, g_x^{\mathbb{H}_K})$:
\begin{equation}
\mathbb{H}_K^n = \{x \in \mathbb{R}^{n+1} : \langle x, x \rangle_\mathcal{L} = 1/K, x_0 > 0 \}.
\end{equation} 
The metric tensor is $g_x^{\mathbb{H}_K} = \operatorname{diag}([-1, 1, \ldots 1])$, and the origin of the hyperboloid model is $\mathbf{o} = (1 / \sqrt{|K|}, 0, \ldots , 0) \in \mathbb{R}^{n+1}$.
The distance between two points $x,y \in \mathbb{H}_K^n$ is defined as
\begin{equation}
d_{\mathbb{H}_K^n}(x,y)=\frac{1}{\sqrt{-K}}\operatorname{arcosh}(K \langle x , y \rangle_{\mathcal{L}}).
\label{dist_lorentz}
\end{equation}
For points $x \in \mathbb{H}_K^n$, tangent vector $v \in \mathcal{T}_x \mathbb{H}_K^n$, and $y \neq \textbf{0}$, $\exp_x : \mathcal{T}_x \mathbb{H}_K^n \rightarrow \mathbb{H}_K^n$ and $\log_x : \mathbb{H}_K^n \rightarrow \mathcal{T}_x \mathbb{H}_K^n$ are defined as
\begin{align}
& \exp_x^K(v)= \cosh (s)x + \sinh (s)\frac{v}{s}, \\
& \log_x^K(y)=\frac{\operatorname{arcosh}(K\langle x,y \rangle_{\mathcal{L}})}{\sqrt{K^2 \langle x,y \rangle_{\mathcal{L}}^2-1}}(y-K \langle x,y \rangle_{\mathcal{L}}x),
\end{align}
where $s = \sqrt{-K}\|v\|_\mathcal{L}$ and $\|x\|_\mathcal{L} = \sqrt{\langle x, x \rangle_\mathcal{L}}$.

\section{Datasets}
\subsection{Network Datasets}
\begin{figure*}
    \includegraphics[trim={0cm 0cm 0cm 0cm}, clip, width=.97\textwidth]{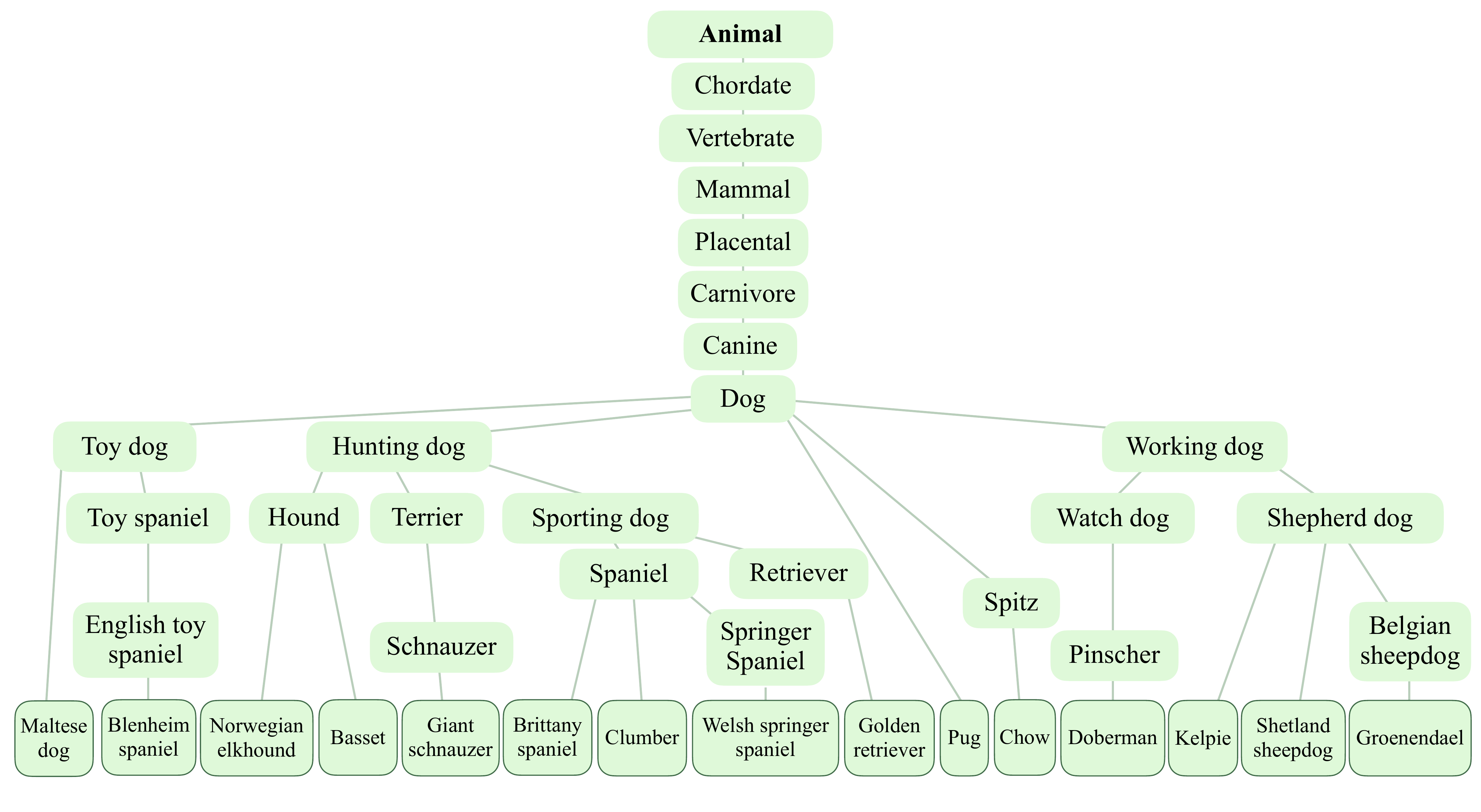}
    \vspace{-5mm}
    \captionof{figure}{Class hierarchy of ImageNet-Dogs\protect\footnotemark.}
    \label{dog}
\end{figure*}
Phylogenetic tree \cite{hofbauer2016preliminary, sanderson1994treebase} models the generic heritage.
CS PhDs \cite{de2018exploratory} represents the relationship between Ph.D. candidates and their advisors in computer science fields.
Diseases \cite{goh2007human, nr-aaai15} is a biological network expressing the relationship between diseases.
Cora \cite{sen2008collective}, Citeseer \cite{sen2008collective}, Pubmed \cite{sen2008collective}, and Wiki \cite{yang2015network} are citation networks whose nodes are scientific papers or web pages and edges represent citation relationships between any two papers or links between any two web pages.
BlogCatalog \cite{tang2009relational} models a social network among bloggers in the online community.
Attribute and label of a node represent the description of each blog and the interest of a blogger, respectively.
Amazon Photo \cite{mcauley2015image} is a part of Amazon co-purchase networks whose nodes are goods and edges represent purchase correlations between any two goods.
A node attribute indicates the bag-of-words for goods' reviews and its label denotes a product category.
\footnotetext{\url{http://image-net.org/index}}

\subsection{Image Datasets}
ImageNet-10 \cite{chang2017deep} and ImageNet-Dogs \cite{chang2017deep} are subsets of the ImageNet dataset \cite{krizhevsky2012imagenet}. 
ImageNet-10 consists of $13,000$ images from $10$ randomly selected subjects.
ImageNet-Dogs are $19,500$ images from $15$ randomly selected dog breeds.
The class hierarchy of ImageNet-Dogs is illustrated in Fig. \ref{dog}. 
We have constructed a new dataset, ImageNet-BNCR, via randomly choosing $3$ leaf classes per root.
We chose three roots, \textit{Artifacts}, \textit{Natural objects}, and \textit{Animal}.
Thus, there exist $9$ leaf classes, and each leaf class contains $1,300$ images in ImageNet-BNCR dataset.
For every dataset used for the image clustering task, we used only the training set without the validation set, and images were resized to $96 \times 96 \times 3$.

\section{Compared Methods}
\subsection{Node Clustering and Link Prediction} We compared HGCAE with seven state-of-the-art unsupervised message passing models which mainly conduct in Euclidean space.
\begin{itemize}[leftmargin=*]
    \item \textbf{GAE \cite{kipf2016variational}, VGAE \cite{kipf2016variational}, ARGA \cite{pan2018adversarially},} and \textbf{ARVGA \cite{pan2018adversarially}} are graph auto-encoders that reconstruct only the affinity matrix using a non-parametric decoder which is not learnable.
    \item \textbf{MGAE \cite{wang2017mgae}} is a stacked one-layer graph auto-encoder that reconstructs only the node attributes via a linear activation function.
    \item \textbf{GALA \cite{park2019symmetric}} is a graph auto-encoder that reconstructs only the node attributes through learnable parametric encoder and decoder.
    \item \textbf{DBGAN \cite{zheng2020distribution}} is a distribution-induced bidirectional generative adversarial network that estimates the structure-aware prior distribution of the representations.
\end{itemize}

GAE \cite{kipf2016variational}, VGAE \cite{kipf2016variational}, ARGA \cite{pan2018adversarially}, ARVGA \cite{pan2018adversarially}, and GALA \cite{park2019symmetric} are constrained to have two-layer auto-encoder models, since they report that two-layer structures show the best performances.
In the case of MGAE \cite{wang2017mgae} which is a stacked one-layer auto-encoder model, we have stacked the layer up to three and reported the best performances.
For DBGAN \cite{zheng2020distribution}, we followed the number of layers in the literature.
For every compared method, we followed the hyperparameters in the literature.

\subsection{Image Clustering}
Extensive baselines and state-of-the-art image clustering methods were compared.
Several traditional methods including k-means clustering (Kmeans) \cite{lloyd1982least}, spectral clustering (SC) \cite{zelnik2005self}, agglomerative clustering (AC) \cite{gowda1978agglomerative}, and nonnegative matrix factorization (NMF) \cite{cai2009locality} were also compared.
For the representation-based clustering methods, AE \cite{bengio2007greedy}, CAE \cite{masci2011stacked}, SAE \cite{ng2011sparse}, DAE \cite{vincent2010stacked}, DCGAN \cite{radford2015unsupervised}, DeCNN \cite{zeiler2010deconvolutional}, SWWAE \cite{zhao2015stacked}, and VAE \cite{kingma2013auto} were adopted.
Besides, the state-of-the-art image clustering methods including JULE \cite{yang2016joint}, DEC \cite{xie2016unsupervised}, DAC \cite{chang2017deep}, DDC \cite{chang2019deep}, DCCM \cite{wu2019deep}, and PICA \cite{huang2020deep} were employed.
For every compared method, we followed the experimental details in the literature.

\section{Experimental Details}
For every experiment and analysis, HGCAE has two encoder layers and two decoder layers.
The dimension of each layer for HGCAE was set to one of $\{2^{3}, 2^{4}, ..., 2^{11} \}$.
We optimized HGCAE using Adam \cite{kingma2015adam} with learning rate $0.01$.
As reported in \cite{chami2019hyperbolic}, we observe that Euclidean optimization \cite{kingma2015adam} is much more stable than Riemannian optimization \cite{becigneul2018riemannian}.
Because of exponential and logarithmic maps, the parameters of our model can be optimized using Euclidean optimization.
We experimented with HGCAE for two cases, fixing the curvature of all layers or learning the curvature of each layer, then we reported the best performances.
In the case of fixing the curvature of all layers, the curvature $K$ was set to one of $\{-0.1, -0.5, -1, -2\}$.
The regularization parameter $\lambda$ of Eq. (12) in the manuscript was set to one of $\{10^{-6}, 10^{-5}, ..., 10^3 \}$.
The initial values of trainable parameters $\beta$ and $\gamma$ in Eq. (9) in the manuscript were set to $0$.
We searched the best hyperparameters which suited well to each dataset by random search.
For visual datasets, we construct the mutual $k$ nearest neighbors graph, $A$, as follows:
\begin{equation}
A_{ij}=\begin{cases}
   1 & \text{if $x_i \in \operatorname{NN}_k(x_j) \wedge x_j \in \operatorname{NN}_k(x_i)$ }
   \\
   0 & \text{otherwise},
\end{cases}
\end{equation}
where $x_i$ and $\operatorname{NN}_k(x_i)$ denote the feature and $k$ Euclidean nearest neighbor set of the $i$-th image respectively.
We set $k=20$ and $k=10$ for ImageNet-10 and ImageNet-Dogs, respectively.
\begin{table*}[t]
\caption{Ablation studies on link prediction task: The baseline model is GAE which conducts graph convolution in Euclidean space, does not use an attention mechanism and reconstructs only the graph structure $A$.} 
\vspace{-5mm}
\begin{center}
\footnotesize
\begin{tabular}{ccccccccccccccc|cc|cc}
\midrule
\multicolumn{3}{c}{\multirow{2}{*}{}} & \multicolumn{3}{c}{Reconstruct} & \multicolumn{3}{c}{Geometry} & \multicolumn{3}{c}{in hyperbolic space} & \multicolumn{3}{c|}{in hyperbolic spaces} & \multicolumn{2}{c|}{Cora} & \multicolumn{2}{c}{Citeseer}\\
& & & \multicolumn{3}{c}{both $A$ and $X$} & \multicolumn{3}{c}{aware attention} & \multicolumn{3}{c}{fixing $K$} & \multicolumn{3}{c|}{learning $K$} & AUC & AP & AUC & AP \\
\midrule        
\multicolumn{3}{l}{Baseline: GAE \cite{kipf2016variational}} & \multicolumn{3}{c}{$\times$} & \multicolumn{3}{c}{$\times$} & \multicolumn{3}{c}{$\times$} & \multicolumn{3}{c|}{$\times$} & 91.0 & 92.0 & 89.5 & 89.9 \\
\multicolumn{3}{l}{Ablation I} & \multicolumn{3}{c}{$\surd$} & \multicolumn{3}{c}{$\times$} & \multicolumn{3}{c}{$\times$} & \multicolumn{3}{c|}{$\times$} & 92.7 & 92.1 & 94.0 & 94.8 \\
\multicolumn{3}{l}{Ablation II} & \multicolumn{3}{c}{$\surd$} & \multicolumn{3}{c}{$\times$} &  \multicolumn{3}{c}{$\surd$} & \multicolumn{3}{c|}{$\times$} & 94.6 & 94.4 & 95.9 & 96.3 \\
\multicolumn{3}{l}{Ablation III} & \multicolumn{3}{c}{$\times$} & \multicolumn{3}{c}{$\surd$} &  \multicolumn{3}{c}{$\surd$} & \multicolumn{3}{c|}{$\times$} & 94.5 & 94.8 & 96.1 & 96.4 \\
\midrule
\multicolumn{3}{l}{\textbf{Proposed I: HGCAE}} & \multicolumn{3}{c}{$\surd$} & \multicolumn{3}{c}{$\surd$} &  \multicolumn{3}{c}{$\surd$} & \multicolumn{3}{c|}{$\times$} & 95.4 & \textbf{95.5} & \textbf{96.7} & \textbf{97.0} \\
\multicolumn{3}{l}{\textbf{Proposed II: HGCAE}} & \multicolumn{3}{c}{$\surd$} & \multicolumn{3}{c}{$\surd$} &  \multicolumn{3}{c}{$\times$} & \multicolumn{3}{c|}{$\surd$} & \textbf{95.6} & \textbf{95.5} & 96.5 & 96.8 \\
\midrule
\end{tabular}
\end{center}
\vspace{-5mm}
\label{ablation1}
\end{table*}

\begin{table}[t]
\caption{Clustering performances in low-dimensional space.}
\vspace{-5mm}
\begin{center}
\scriptsize
\begin{tabularx}{0.45\textwidth}{lYYYYYY}
\midrule
\multirow{3}{*}{} & \multicolumn{2}{c}{Pubmed} & \multicolumn{2}{c}{BlogCatalog} & \multicolumn{2}{c}{Amazon Photo} \\
\cmidrule{2-7}
& ACC & NMI & ACC & NMI & ACC & NMI \\
\midrule        
GAE   \cite{kipf2016variational}           & 51.3 & 7.7 & 27.6  & 11.4 & 37.1 & 27.3 \\
VGAE  \cite{kipf2016variational}           & 40.6 & 0.1 & 23.3  & 5.9  & 36.3 & 27.7 \\
ARGA  \cite{pan2018adversarially}          & 40.0 & 0.5 & 29.8  & 14.6 & 41.0 & 37.0 \\
ARVGA \cite{pan2018adversarially}          & 38.5 & 0.1 & 27.2  & 9.7  & 40.8 & 27.8 \\
GALA  \cite{park2019symmetric}             & 36.1 & 0.4 & 25.2  & 7.1  & 24.2 & 5.8  \\
\midrule
\textbf{HGCAE}               & \textbf{68.1}& \textbf{28.2} & \textbf{74.1} & \textbf{57.8} & \textbf{76.3} & \textbf{64.0} \\
\midrule
\end{tabularx}
\label{ablation2}
\end{center}
\vspace{-6mm}
\end{table}

\subsection{Details of Node Clustering and Link Prediction}
For the link prediction task, we divided the edges into training edges, validation edges, and test edges as $85\%$, $5\%$, and $10\%$, then we used validation edges for the model convergence. 
During training for the link prediction task, we only reconstructed training edges in $\mathcal{L}_{REC-A} = \mathbb{E}_{q(H|X, A)}[\log p(\hat{A}|H)]$.
For the node clustering task, every edge is reconstructed by the output of the encoder during training.
The performance of node clustering was obtained by running k-means clustering \cite{lloyd1982least} on the latent representations (output of the encoder) in the tangent space of the last layer of the encoder.

\subsection{Details of Image Clustering}
The performance of HGCAE on the image clustering task was obtained by running k-means clustering \cite{lloyd1982least} on the latent representations (output of the encoder) in the tangent space of the last layer of the encoder.

\subsection{Details of Convolutional Auto-Encoder}
We extracted $1000$-dimensional features by training a convolutional auto-encoder (CAE) \cite{masci2011stacked} on the ImageNet-10 \cite{chang2017deep} and ImageNet-BNCR datasets on the experiment of Section 5.3 in the manuscript.
We used the encoder part and decoder part as VGG-16 network \cite{simonyan2014very} and five deconvolution layers \cite{zeiler2010deconvolutional} respectively.
We optimized CAE using Adam \cite{kingma2015adam} with learning rate $0.0001$ and obtained the feature after $100$ epochs.

\subsection{Details of Image Classification}
We obtained the latent representation of ImageNet-10 \cite{chang2017deep} and ImageNet-BNCR by training CAE on the experiments of Section 5.4 in the manuscript.
For the image classification task, we trained the VGG-11 \cite{simonyan2014very} classifier.
We trained the classifier using stochastic gradient descent \cite{bottou1998online} and used the learning rate scheduler as in \cite{yun2019cutmix}.
When adding further samples in every training epoch, high, middle, and low HDO samples were chosen by $n \%$ of the original data closest to the boundary, $n\%$ of the original data closest to the median of distance histogram, and $n\%$ of the original data closest to the origin, respectively.
We set $n$ for ImageNet-10 and ImageNet-BNCR to $30$ and $50$ respectively.
The learning rates of ImageNet-10 and ImageNet-BCNR were set to $0.01$ and $0.0005$ respectively.
When training BaselineFL, we tried $\{0.5, 1.0, 2.0\}$ for $\gamma$ in focal loss \cite{lin2017focal} and reported the best performances.
There has been recent research on manipulating the gradient updates based on the prediction difficulty, anchor loss (AL) \cite{ryou2019anchor}, and we have tried to report the classification performance of AL as well as FL.
However, due to the several NaN issues of official AL implementation\footnote{\url{https://github.com/slryou41/AnchorLoss}}, we could not report the performance of AL.

\section{Further Experiments}
\subsection{Effectiveness of The Proposed Components}
Through link prediction experiments, we validated the effectiveness of two components: learning in the hyperbolic spaces and reconstructing both the graph structure and the node attributes.
The experiment was conducted on two citation networks, Cora \cite{sen2008collective} and Citeseer \cite{sen2008collective}, then the results for link prediction task are presented in Table \ref{ablation1}.
The baseline model is GAE \cite{kipf2016variational}, which conducts graph convolution in Euclidean space, does not use an attention mechanism, and reconstructs only the affinity matrix $A$.
In Ablation I, reconstructing both the node attribute $X^{Euc}$ and the graph structure $A$ (Eq. (12) in the manuscript) are added to the baseline settings. 
In Ablation II, operating in hyperbolic space with fixed curvature $K$ is added to Ablation I. 
In Ablation III, operating in hyperbolic space with fixed curvature $K$ and the geometry-aware attention mechanism (Eq. (9) in the manuscript) are added to baseline settings.
The results between Ablation I and Ablation II show that the message passing in the hyperbolic space is more effective than that in Euclidean space.  
Also, the performance gap between Ablation III and Proposed I shows that it is helpful to learn a representation that reflects both the structure of the network and the attributes of each node in hyperbolic space.
This component is also valid in Euclidean space, as shown in the gap between Baseline and Ablation I. 
As shown in the gap between Proposed I and II, the fixed $K$ and the trainable $K$ show similar performance to each other for some datasets, but training $K$ gives an efficient training scheme without multiple learning for searching the best $K$.

\subsection{Learning in Low-Dimensional Space} 
\begin{figure}[t]
\centering
\includegraphics[trim={0.4cm 28.5cm 9cm 0cm}, clip, width=.49\textwidth]{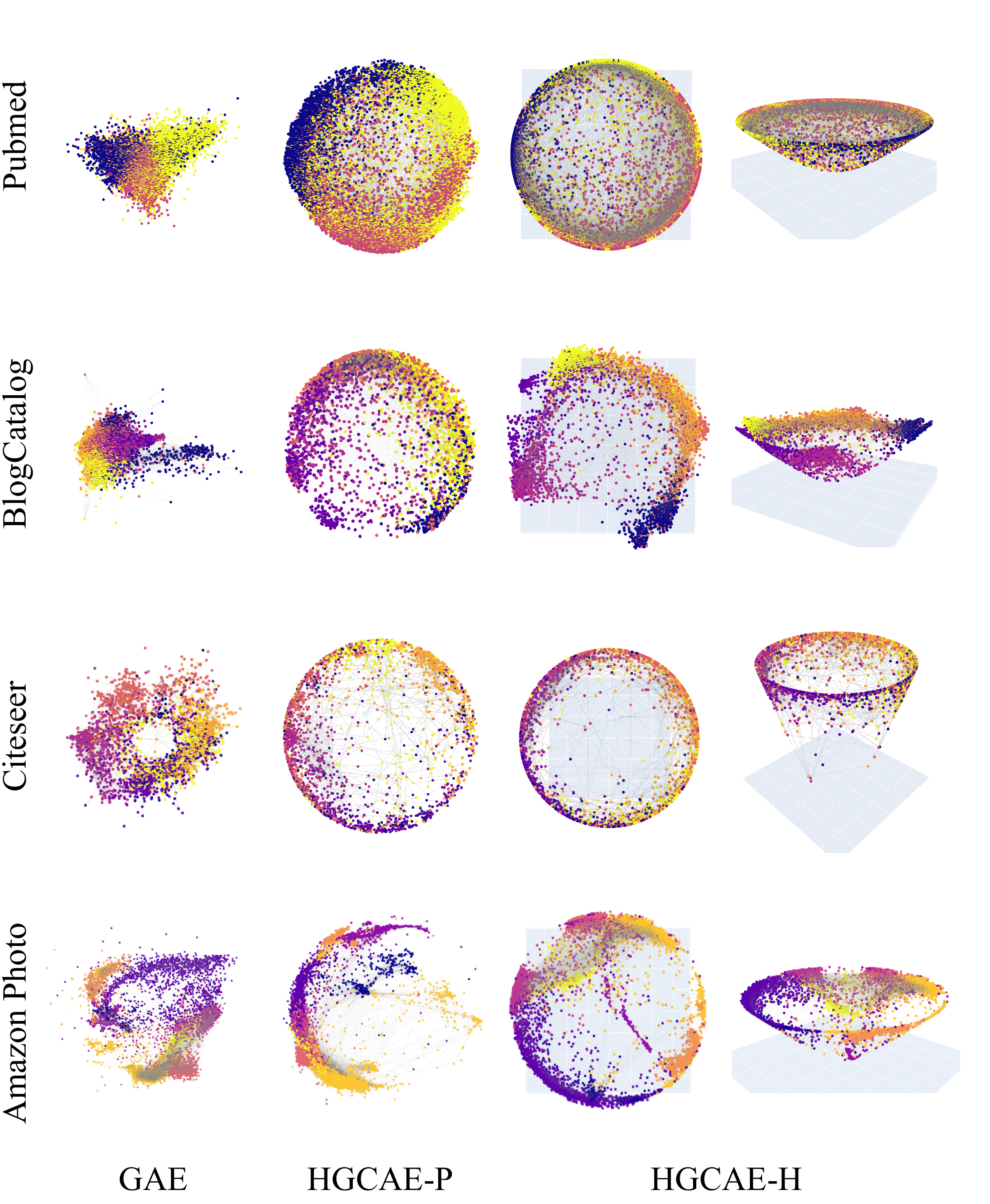}
\caption{2-dimensional embeddings in Euclidean, Poincar\'e ball, and hyperboloid latent spaces on Pubmed, BlogCatalog, Citeseer, and Amazon Photo datasets.}
\label{network_embedding}
\vspace{-3mm}
\end{figure}
One of the strengths of hyperbolic space compared to Euclidean space is that hyperbolic model can learn latent representation of data whose structure is hierarchical without the need for infeasible high-dimensional space \cite{de2018representation}.
To show this point, we obtained the latent representations of network datasets in the very low-dimensional latent space for node clustering task.
Every compared graph auto-encoder and HGCAE were constrained to have two layers whose each dimension was $4$ and $2$ respectively.
Note that the performance of MGAE \cite{wang2017mgae} cannot be reported since MGAE cannot manipulate the latent dimension.
The experiments were conducted on Pubmed \cite{sen2008collective}, BlogCatalog \cite{tang2009relational}, and Amazon Photo \cite{mcauley2015image} datasets. 
The results are presented in Table \ref{ablation2}.
Although the dimension of latent space is extremely low, HGCAE still significantly outperforms the state-of-the-art unsupervised message passing methods operating in Euclidean space.
Notably, on BlogCatalog and Amazon Photo datasets, HGCAE achieves more than $30\%$ higher performances compared to Euclidean counterparts.
These results support that hyperbolic space is effective than Euclidean space even in the very low-dimensional latent space.

\subsection{Visualization of The Network Datasets} We explored the latent representations of GAE \cite{kipf2016variational} and our models on Pubmed \cite{sen2008collective}, BlogCatalog \cite{tang2009relational}, Citesser \cite{sen2008collective}, and Amazon Photo \cite{mcauley2015image} datasets by constraining the latent space as a $2$-dimensional hyperbolic or Euclidean space.
The result is given in Fig. \ref{network_embedding}.
On the results of HGCAE, most of the nodes are located on the boundary of hyperbolic space and well-clustered with the nodes in the same class.

\subsection{Sensitivity of Hyperparameter Setting}
One of the important hyperparameters of HGCAE is $\lambda$ in Eq. (12) in the manuscript.
If $\lambda$ is required large (small) value, this means that the node attributes (subgraph structures) are the more important factor of latent representation. 
Since node attributes and the graph structure are different for each dataset, the optimal $\lambda$ has different values for each dataset.
In cases of BlogCatalog and Citeseer (Cora), we empirically found that small (large) $\lambda$ value is optimal for both link prediction and node clustering tasks.

\section{Further Discussions}
\subsection{Connection to Contrastive Learning}
The hyperbolic geometry can be extended to contrastive learning \cite{chen2020simple}.
A recent study \cite{tschannen2019mutual} has uncovered the link between contrastive learning and deep metric learning.
In this respect, it is becoming more significant to find the informative (hard) negative samples, embeddings that are difficult to distinguish from anchors, beyond uniform sampling \cite{robinson2021contrastive}.
Our work empirically showed that \textit{H}yperbolic \textit{D}istance from the \textit{O}rigin (HDO) is an effective criterion for selecting samples without supervision for better generalization.
The concept of HDO could be extended to informative negative sampling.
Since the embeddings hard to discriminate is equal to those that are hard to classify by the model, the samples near the origin of hyperbolic space can be the impactful negative samples to increase the ability of the unsupervised contrastive learning.

\subsection{Failure Cases of Hyperbolic Embedding Spaces}
The inductive bias of hyperbolic representation learning is assuming that there exist hierarchical relationships in the dataset.
Thus if the structure of the graph modeling the relation between data points is close to a tree, the hyperbolic space, a continuous version of a tree, is a suitable latent space. 
However, not all datasets' latent structures have the topological properties of the tree.
For instance, datasets obtained from omnidirectional sensors of drones and autonomous cars are indeed more suitable to latent hyperspherical manifold rather than the hyperbolic manifold \cite{cohen2018spherical}.

\end{document}